%% file: main_arxiv.tex
\documentclass[sigconf]{acmart}

\AtBeginDocument{%
  }

\input{_marcros}

\newcommand\blfootnote[1]{%
  \begingroup
  \renewcommand\thefootnote{}\footnote{#1}%
  \addtocounter{footnote}{-1}%
  \endgroup
}

\setcopyright{acmlicensed}
\copyrightyear{2025}
\acmYear{2025}
\acmConference[MM '25]{Proceedings of the 33rd ACM International Conference on Multimedia}{October 27--31, 2025}{Dublin, Ireland}
\acmBooktitle{Proceedings of the 33rd ACM International Conference on Multimedia (MM '25), October 27--31, 2025, Dublin, Ireland}
\acmISBN{979-8-4007-2035-2/2025/10}
\acmDOI{10.1145/3746027.3755081}

\begin{document}

%%
%% The "title" command has an optional parameter,
%% allowing the author to define a "short title" to be used in page headers.
\title{Individual Content and Motion Dynamics Preserved Pruning for Video Diffusion Models}

%%
%% The "author" command and its associated commands are used to define
%% the authors and their affiliations.
%% Of note is the shared affiliation of the first two authors, and the
%% "authornote" and "authornotemark" commands
%% used to denote shared contribution to the research.
% \author{Ben Trovato}
% \authornote{Both authors contributed equally to this research.}
% \email{trovato@corporation.com}
% \orcid{1234-5678-9012}
% \author{G.K.M. Tobin}
% \authornotemark[1]
% \email{webmaster@marysville-ohio.com}
% \affiliation{%
%   \institution{Institute for Clarity in Documentation}
%   \city{Dublin}
%   \state{Ohio}
%   \country{USA}
% }
\settopmatter{authorsperrow=4}
\author{Yiming Wu}
\affiliation{%
  % \institution{School of Computing and Data Science, The University of Hong Kong}
  \institution{\hspace{-3mm} The University of Hong Kong}
  \city{Hong Kong}
  \country{China}
}
% \email{yimingwu@hku.hk}

\author{Zhenghao Chen*}
\affiliation{%
  % \institution{School of Information and Physical Sciences, The University of Newcastle}
  \institution{\hspace{-1.5mm} University of Newcastle}
  \city{Newcastle}
  \country{Australia}
}
% \email{zhenghao.chen@newcastle.edu.au}

\author{Huan Wang}
\affiliation{%
  % \institution{School of Engineering, Westlake University}
  \institution{\hspace{-3.5mm}  Westlake University}
  \city{Hangzhou}
  \country{China}
}
% \email{wanghuan@westlake.edu.cn}

\author{Dong Xu*}
\affiliation{%
  % \institution{School of Computing and Data Science, The University of Hong Kong}
  \institution{\hspace{-3.5mm}The University of Hong Kong}
  \city{Hong Kong}
  \country{China}
}
% \email{dongxu@hku.hk}

% \begin{verbatim}
% \authorsaddresses{Corresponding author: Ben Trovato,
% \href{mailto:trovato@corporation.com}{trovato@corporation.com};
% Institute for Clarity in Documentation, P.O. Box 1212, Dublin,
% Ohio, USA, 43017-6221} 
% \end{verbatim}

% \cs{authorsaddresses}
% \begin{verbatim}
% \authorsaddresses{Corresponding author: Zhenghao Chen and Dong Xu}
% \end{verbatim}

%%
%% By default, the full list of authors will be used in the page
%% headers. Often, this list is too long, and will overlap
%% other information printed in the page headers. This command allows
%% the author to define a more concise list
%% of authors' names for this purpose.
\renewcommand{\shortauthors}{Wu et al.}

%%
%% The abstract is a short summary of the work to be presented in the
%% article.
\begin{abstract}
  The high computational cost and slow inference time are major obstacles to deploying Video Diffusion Models (VDMs). To overcome this, we introduce a new Video Diffusion Model Compression approach using individual content and motion dynamics preserved pruning and consistency loss.
  First, we empirically observe that deeper VDM layers are crucial for maintaining the quality of \textbf{motion dynamics} (\textit{e.g.,} coherence of the entire video), while shallower layers are more focused on \textbf{individual content} (\textit{e.g.,} individual frames). Therefore, we prune redundant blocks from the shallower layers while preserving more of the deeper layers, resulting in a lightweight VDM variant called VDMini.
  Moreover, we propose an \textbf{Individual Content and Motion Dynamics (ICMD)} Consistency Loss to gain comparable generation performance as larger VDM to VDMini. In particular, we first use the Individual Content Distillation (ICD) Loss to preserve the consistency in the features of each generated frame between the teacher and student models. Next, we introduce a Multi-frame Content Adversarial (MCA) Loss to enhance the motion dynamics across the generated video as a whole. This method significantly accelerates inference time while maintaining high-quality video generation.
  Extensive experiments demonstrate the effectiveness of our VDMini on two important video generation tasks, Text-to-Video (T2V) and Image-to-Video (I2V), where we respectively achieve an average 2.5 $\times$, 1.4 $\times$, and 1.25 $\times$ speed up for the I2V method SF-V, the T2V method T2V-Turbo-v2, and the T2V method HunyuanVideo, while maintaining the quality of the generated videos on several benchmarks including UCF101, VBench-T2V, and VBench-I2V.
\end{abstract}

%%
%% The code below is generated by the tool at http://dl.acm.org/ccs.cfm.
%% Please copy and paste the code instead of the example below.
%%



%% Keywords. The author(s) should pick words that accurately describe
%% the work being presented. Separate the keywords with commas.
%% A "teaser" image appears between the author and affiliation
%% information and the body of the document, and typically spans the
%% page.
% \begin{teaserfigure}
%   \includegraphics[width=\textwidth]{sampleteaser}
%   \caption{Seattle Mariners at Spring Training, 2010.}
%   \Description{Enjoying the baseball game from the third-base
%   seats. Ichiro Suzuki preparing to bat.}
%   \label{fig:teaser}
% \end{teaserfigure}

% \received{20 February 2007}
% \received[revised]{12 March 2009}
% \received[accepted]{5 June 2009}

%%
%% This command processes the author and affiliation and title
%% information and builds the first part of the formatted document.
\maketitle

\input{01_intro}
\input{02_related}
\input{03_method}
\input{04_experiments}

\input{10_conclusion}

%%
%% The acknowledgments section is defined using the "acks" environment
%% (and NOT an unnumbered section). This ensures the proper
%% identification of the section in the article metadata, and the
%% consistent spelling of the heading.
% \begin{acks}
% The research work described in this paper was conducted in the JC STEM Lab of Multimedia and Machine Learning funded by The Hong Kong Jockey Club Charities Trust.
% \end{acks}

%%
%% The next two lines define the bibliography style to be used, and
%% the bibliography file.
\bibliographystyle{ACM-Reference-Format}
\balance
\bibliography{sample-base}

%%
%% If your work has an appendix, this is the place to put it.
\appendix

\input{12_appendix}

\end{document}

%% file: _marcros.tex
\usepackage{subfig}
\usepackage{graphicx}
\usepackage{animate}
\usepackage{tcolorbox}
\usepackage{tabularx}
\usepackage{multirow}
\usepackage{array}
\usepackage[normalem]{ulem}

% ...

\newcommand{\ie}[0]{\emph{i.e.,}}
\newcommand{\etal}[0]{\emph{et al}}
\newcommand{\nbf}[1]{{\noindent \textbf{#1.}}}

\newcommand{\red}[1]{{\color{red}#1}}

\newcommand{\numColumns}{4}
\newcommand{\columnSpacing}{0.15em}
\newcommand{\animationNotes}{
\textit{Best viewed with Acrobat Reader for animation. Click the last image to play the animation clips.}
}

%% file: 01_intro.tex
\section{Introduction}
\label{sec:intro}

% To insert a figure: \input{figs/template}
% Or table: \input{tables/template}
% background about video generation
Video generation has achieved significant progress in the recent few years, owing to the rapid development of the video diffusion model (VDM)~\cite{song2024improved,songConsistencyModels2023,blattmannStableVideoDiffusion2023}. % and autoregressive models~\cite{ondratyukVideoPoetLargeLanguage2024}. 
Despite the promising advances in video diffusion quality, the high computational costs and slow inference time hinder the democratization of video generation tasks in real-world applications.
For instance, generating a two-second video by Stable Video Diffusion~\cite{blattmannStableVideoDiffusion2023} costs more than 25 seconds on an A100 GPU, and generating a 5 seconds video costs more than 5 minutes on online platforms such as KLing~\cite{kling} and DreamMachine~\cite{luma}.
\blfootnote{
This work was done during Yiming Wu's fellowship at The University of Hong Kong
$^*$Zhenghao Chen and Dong Xu are the corresponding authors. \\ 
\textit{E-mail address: \{yimingwu, dongxu\}@hku.hk, zhenghao.chen@newcastle.edu.au and wanghuan@westlake.edu.cn}}

% current sota video diffusion models
% To overcome the computational bottleneck of such video diffusion model, the straightforward solutions are reduce sampling steps or model parameters. Recent approaches focus more on sampling steps reduction. For example, some works intuitively extends consistency distillation~\cite{songConsistencyModels2023,wangAnimateLCMAcceleratingAnimation2024,zhaiMotionConsistencyModel2024} and others adopted adversarial training~\cite{zhangSFVSingleForward2024,maoOSVOneStep2024} to VDMs.

To address the computational challenges in VDM, two primary strategies can be adopted: reducing the number of sampling steps and decreasing model parameters.
Recent research has predominantly focused on minimizing sampling steps. For instance, some works have extended consistency distillation techniques~\cite{songConsistencyModels2023,wangAnimateLCMAcceleratingAnimation2024,zhaiMotionConsistencyModel2024, chen2022lsvc}, while other approaches have employed adversarial training methods~\cite{zhangSFVSingleForward2024,maoOSVOneStep2024} to VDM.
On the other hand, model compression techniques for VDM remain under-explored. While the common model compression techniques such as pruning~\cite{reed1993pruning,hoefler2021sparsity,han2015learning,li2017pruning,wang2021neural,fang2023depgraph,chen2022exploiting} and knowledge distillation~\cite{bucilua2006model,hinton2015distilling} have shown promising results on image diffusion models~\cite{liSnapFusionTextImageDiffusion2023,kimBKSDMLightweightFast2023,castellsEdgeFusionDeviceTextImage2024}, directly applying these methods to video diffusion models is still challenging. 
% For model compression, the different blocks of VDM focus on the different content in video generation, which leads to a content-aware pruning technique.
% And for fine-tuning the pruned model, 

%
% challenges:
% 1.  different blocks perform different roles in image diffusion and video diffusion ( talk about your pruning motivation), shallow for XXX, deeper for XXX, so we propose VDMini
% 2.  video content has individual and integrate focus, so we use individual and integrate loss XXX
%
% First, different from the image diffusion model, where the importance of blocks can be simply estimated by evaluating the quality of a single generated image, the VDM pruning needs to assess the quality of generated contents from individual frames and the dynamic of integrated generated video. After a comprehensive assessment, we observe that deeper VDM layers play a vital role in preserving the quality of integrated content, such as motion dynamics for overall video, whereas shallower layers are primarily dedicated to individual generated RGB content from each single frame.

First, unlike the image diffusion models, where block importance can be relatively easily estimated by assessing the quality of a single generated image, pruning in VDMs requires a more complex evaluation. In VDMs, it's essential to assess both the quality of content in individual frames and the motion dynamic consistency across the entire generated video~\cite{chen2024group, chen2023neural}.
Through comprehensive analysis, we have observed that the deeper layers of VDMs are crucial for maintaining multi-frame content consistency, particularly in terms of motion dynamics across frames. In contrast, the shallower layers primarily contribute to the generation of individual RGB content in each frame. This insight allows us to selectively prune shallower layers while preserving the deeper ones, which optimizes computational efficiency without compromising video quality.
% the shallow blocks are critical for individual content consistency (actually FVD score, see Fig. 2(a)), and the deeper layers are important for preserving motion dynamics since we observe the ``video frozen'' (replace it) when removing the deeper blocks in the downblocks and upblocks of U-Net (Downblock.2 and Upblock.1, not the innermost blocks, i.e. Downblock.3, Midblock, and Upblock.0). 
With such observation, we present a light-weight VDM architecture \textbf{VDMini}, which achieves a 2.5$\times$ and 1.4$\times$ speedup for the image-to-video (I2V) and the text-to-video (T2V) generation tasks, respectively.
% reduces the unet of original VDMs i.e. SF-V and T2V-Turbo-v2 by 40\% and 43\%

% % consistency loss
% 2. recovering video content from noises is more challenging than images, which requires the maintenance of spatial smoothness in the individual frames as well as the coherence of content (semantic consistency) across videos. Thus, we adopt the individual content loss to transfer the knowledge from the teacher model to the student model in the multiple feature levels. Then we introduce the integrated adversarial loss to enhance the integration of the generated video, where a min-max game is set up between the discriminator with spatial and temporal head and student model. Specifically, the discriminator is trained to distinguish the generated contents (actually the extracted latents) from the frozen teacher model and student model, and the student model aims to fool the discriminator (generate the video content as close as the real samples).

To further enhance the performance of the pruned VDM (i.e., VDMini) and achieve comparable generation quality to the unpruned VDM, we perform a post-pruning fine-tuning procedure. Given the aforementioned challenges, we focus on maintaining both individual content quality (i.e., spatial smoothness within a single frame) and motion dynamics quality (i.e., coherence across the entire video). To this end, we propose a novel Individual Content and Motion Dynamics (ICMD) Consistency Loss, which ensures consistency between the pruned VDMini (i.e., student) and the unpruned VDM (i.e., teacher). Drawing inspiration from prior work in image generation~\cite{kimBKSDMLightweightFast2023}, we employ knowledge distillation to preserve individual content consistency. Additionally, to ensure multi-frame content consistency without redundancy, we introduce an adversarial loss on video-level feature representations, reinforcing overall content coherence.

% and adversarial training to transfer knowledge from a larger VDM (teacher) to a smaller VDM (student). Specifically, Individual Content Distillation (ICD) Loss ensures the consistency of individual content, i.e. individual frames, between the teacher and student models, and Integrated Content Adversarial (ICA) Loss enhances the motion dynamics across the generated video as a whole. 

% Specifically, we first eliminate network blocks according to the block importance and qualitative results, which is crucial for maintaining the quality of integrated content, i.e. motion dynamics. Next, we fine-tune the pruned model with the ICMD loss, which consists of two components: Individual Content Distillation (ICD) Loss and Integrated Content Adversarial (ICA) Loss. The ICD Loss ensures the consistency of individual content, i.e. individual frames, between the teacher and student models, while the ICA Loss enhances the motion dynamics across the generated video as a whole. 

% experiment
% We validate the effectiveness of our approach on two most representative video generation tasks, Image-to-Video (I2V) and Text-to-Video (T2V), where we respectively achieve a 2.5 $\times$ and 1.4 $\times$ speedup for I2V model SFV~\cite{zhangSFVSingleForward2024} and T2V model T2V-Turbo-V2~\cite{liT2VTurbov2EnhancingVideo2024}, while maintaining the comparable generation quality
We validate the effectiveness of our approach on two representative video generation tasks: Image-to-Video (I2V) and Text-to-Video (T2V) on the UCF101 and VBench benchmarks. By combining the proposed block pruning technique with the ICMD loss, our method achieves a 2.5$\times$ speedup on the I2V model SF-V~\cite{zhangSFVSingleForward2024} and a 1.4$\times$ speedup on the T2V model T2V-Turbo-V2~\cite{liT2VTurbov2EnhancingVideo2024}, while maintaining comparable generation quality. Moreover, for T2V tasks, we conducted additional experiments using the latest diffusion transformer-based T2V model, Hunyuan~\cite{kong2024hunyuanvideo}, achieving a 25\% speedup and demonstrating generalizability.
% This demonstrates the potential of our approach to improve efficiency without compromising on the quality of generated videos.
%
% We validate the effectiveness of our approach on two key video generation tasks: Image-to-Video (I2V) and Text-to-Video (T2V). Our method achieves a 2.5$\times$ speedup on the I2V model SF-V~\cite{zhangSFVSingleForward2024} and a 1.4$\times$ speedup on the T2V model T2V-Turbo-V2~\cite{liT2VTurbov2EnhancingVideo2024}, all while preserving comparable generation quality on the UCF101 and VBench benchmarks. These results demonstrate our approach's potential to enhance efficiency without compromising the quality of generated videos.
%
Our contributions are summarized as follows:
\begin{itemize}
\item We conduct a comprehensive assessment for the importance estimation of VDM blocks based on individual content and motion dynamics, followed by pruning the redundant layers to achieve model compression. To this end, we present a lightweight VDM variant VDMini, which achieves a significant speedup in inference time.
\item  To ensure quality consistency between the pruned VDMini and the unpruned VDM, we introduce the ICMD loss, it consists of two components: the ICD Loss for individual content consistency, which aligns the generated content in each single frame, and the MCA Loss for multi-frame content consistency, which preserves motion dynamics across the entire video. By maintaining frame-level and video-wide consistency,  we can align VDMini with the unpruned VDM.
\item We validate the effectiveness of our approach on two video generation tasks, namely I2V and T2V. Our method achieves a 2.5 $\times$ speedup over the I2V model SF-V and a 1.4$\times$(\textit{resp.} 1.25 $\times$) speedup over the T2V model T2V-Turbo-v2 (\textit{resp.} HunyuanVideo) while maintaining comparable quality metrics on the benchmarks UCF101, VBench-T2V, and VBench-I2V. 

\end{itemize} 

%% file: 02_related.tex
\section{Related Work}
\label{sec:related}

\subsection{Video Generation}
\label{sec:related:video_generation}
% Introduce current video generation models, including Stable Video Diffusion, Animatediff, VideoCrafter2, Lattie, Sora, Open-Sora, CogVideo, and others.
Video generation has seen wide exploration in recent years. Traditionally, generative adversarial networks (GANs)~\cite{goodfellow2020generative} and VAEs~\cite{kingma2013auto} have been utilized for this purpose, but they are generally limited to producing low-quality and short-duration videos. The advent of diffusion models and large-scale video datasets has significantly enhanced video generation quality. Initial explorations~\cite{khachatryan2023text2video,zhang2024controlvideo} leverage pre-trained image models for video generation without the need for fine-tuning \ie training-free methods. However, despite the reduction in training requirements, these approaches are limited in terms of both inference speed and generation quality due to their reliance on DDIM Inversion~\cite{song2021denoising}.

With the emergence of large-scale video datasets such as WebVid-10M~\cite{bain2021frozen} and Panda70M~\cite{chen2024panda70m}, training-based methods have demonstrated significant improvements in generation quality and condition consistency capabilities. Jonathan \etal~\cite{ho2022video} were the first to extend diffusion models to 3D space-time modeling by incorporating factorized space-time attention blocks into a U-Net architecture.

To leverage pre-trained image diffusion models, approaches such as Video LDM~\cite{blattmann2023align}, Animatediff~\cite{guo2024animatediff}, VideoCrafter2~\cite{chenVideoCrafter2OvercomingData2024}, and Stable Video Diffusion~\cite{blattmannStableVideoDiffusion2023} have been developed. These methods introduce temporal layers following the residual and attention blocks of image generation models, and then employ multi-stage fine-tuning on high-quality video datasets, leading to substantial improvements in generation quality. 

On the other hand, pure transformer-based models, such as DiT~\cite{Peebles2022DiT}, originally designed for image generation, have also been adapted for video generation. Sora~\cite{sora2024} achieved remarkable results by scaling up transformer-based architectures for video generation. To make advanced video generation models more accessible, Open-Sora~\cite{opensora} and Open-Sora-Plan~\cite{pku_yuan_lab_and_tuzhan_ai_etc_2024_10948109} were introduced as open-source solutions for the community. Additionally, CogVideoX~\cite{yang2024cogvideox} proposed an Expert Transformer model for long-duration video generation, surpassing previous state-of-the-art methods. Recent works such as Snap Video~\cite{menapaceSnapVideoScaled2024}, Mochi-1~\cite{genmo2024mochi}, HunyuanVideo~\cite{kong2024hunyuanvideo}, and Wan2.1~\cite{wan2025} have also demonstrated impressive advancements in video generation tasks. Concurrently, methods like SF-V~\cite{zhangSFVSingleForward2024}, AnimateLCM~\cite{wangAnimateLCMAcceleratingAnimation2024}, OSV~\cite{maoOSVOneStep2024}, and T2V-Turbo~\cite{liT2VTurbov2EnhancingVideo2024} focus on reducing inference steps to as few as one to four, albeit with some trade-offs in video quality.

Our approach aims to accelerate video generation by compressing existing high-cost models while preserving generation quality. Our work addresses two specific video generation tasks: image-to-video (I2V) and text-to-video (T2V), through our proposed lightweight model, VidMini, which achieves both efficiency and quality.

\subsection{Pruning for Diffusion Models}
\label{sec:related:pruning on diffusion models}
% Introduce pruning on diffusion models, and the related model compression techniques, mainly classified into three catergories: network quantization, knowledge distillation, and model pruning. In this paper, we focus on model pruning and knowledge distillation.
The excessive size of diffusion models has prompted research on compressing diffusion models via two classic model compression techniques: network pruning, and distillation. Here we summarize the recent advances on them. 

% SnapFusion~\cite{liSnapFusionTextImageDiffusion2023} is one of the pioneering works that accelerate diffusion models from two aspects: changing the model architecture by pruning~\cite{han2015learning,li2017pruning,wang2021neural,fang2023depgraph}; reducing the inference steps by distillation~\cite{hinton2015distilling,salimans2022progressive,meng2023distillation}. 
% The central problem of pruning is to measure the relative weight importance. SnapFusion introduces an importance score comprising two terms: CLIP score drop and latency reduction. 
% They adopt a widely-used ``trial-and-error'' paradigm~\cite{mozer1988skeletonization,molchanov2017pruning} to obtain the importance score for each weight module, remove a module of the model and record the CLIP score drop and latency reduction; the modules with the \textit{least} CLIP score drop and \textit{most} latency reduction are considered insignificant and thus removed. 
% After pruning, a new more efficient UNet architecture is delivered, which runs over 7$\times$ faster than its SD-v1.5 counterpart. To reduce the inference steps, SnapFusion introduces a CFG-aware distillation loss that aligns the student model's output with its teacher's after applying CFG. Overall, they achieved an unprecedented mobile inference speed of less than 2 seconds on an iPhone 14 Pro with no degradation on CLIP score and FID score.

SnapFusion~\cite{liSnapFusionTextImageDiffusion2023} stands out as an early contributor to accelerating diffusion models, employing two primary strategies: architectural refinement through pruning~\cite{han2015learning,li2017pruning,wang2021neural,fang2023depgraph}, and reducing inference iterations via distillation~\cite{hinton2015distilling,salimans2022progressive,meng2023distillation}. Central to pruning efforts is effectively assessing the relative importance of model weights. To this end, SnapFusion proposes an importance metric based on two critical factors: reduction in CLIP score and improvement in latency. 
Specifically, they employ a prevalent trial-and-error approach~\cite{mozer1988skeletonization,molchanov2017pruning}, systematically removing individual modules to measure associated impacts on CLIP scores and latency. Modules demonstrating minimal CLIP degradation alongside maximal latency improvement are identified as candidates for pruning. This approach yields a significantly streamlined UNet architecture that achieves a performance enhancement exceeding 7-fold compared to the original SD-v1.5 model.

For further inference optimization, SnapFusion introduces a CFG-aware distillation technique, ensuring the student model outputs closely match those of the teacher model post-CFG application. Consequently, SnapFusion attains an exceptional inference speed under 2 seconds on an iPhone 14 Pro, preserving the original model's CLIP and FID scores.

Concurrent to SnapFusion, BK-SDM~\cite{kimBKSDMLightweightFast2023} also attempts to accelerate SD models by pruning weight blocks. Unlike SnapFusion, their block importance analysis only considers CLIP score drop and they do not reduce the inference steps. Importantly, they propose a feature-distillation-based retraining strategy to regain model performance after pruning. Overall, BK-SDM achieves 30\% to 50\% reduction in model size and latency against the original SD model, with a small degradation in CLIP score and FID. The compact BK-SDM-Tiny model is further enhanced to EdgeFusion~\cite{castells2024edgefusion} by leveraging a strong teacher, LCM~\cite{luoLATENTCONSISTENCYMODELS2023} to reduce inference steps, and high-quality AI-generated data.

MobileDiffusion~\cite{zhaoMobileDiffusionSubsecondTextImage2023} is a subsequent work that accelerates SD models by improving its architecture and reducing inference steps, too. Like SnapFusion and BK-SDM, it also utilizes pruning to remove redundant residual blocks. Differently, MobileDiffusion proposes even more fine-grained architecture optimizations, such as using more transformers in the middle of U-Net and fewer channels, decoupling self-attention from cross-attention, sharing key-value projections, using separable convolutions, and so on. For reducing inference steps, they adopt the distillation loss from prior works, SnapFusion~\cite{liSnapFusionTextImageDiffusion2023} and UFOGen~\cite{xu2024ufogen}. Overall, by integrating these optimization techniques, they achieve a remarkable inference speed of 0.2 second on an iPhone 15 Pro. 
SnapGen~\cite{chen2025snapgen}, another recent effort, pushes the boundary further by jointly redesigning the architecture and introducing adversarial-aware distillation strategies to support high-resolution image synthesis on mobile devices. It achieves strong generation quality using as few as 4 to 8 denoising steps while maintaining a lightweight model footprint.
This line of work is extended to the video domain in SnapGen-V~\cite{wu2025snapgen}, a concurrent effort to ours, which introduces spatiotemporal attention mechanisms and temporally-aware distillation to support efficient video generation with competitive fidelity and temporal consistency on resource-constrained platforms.

In addition to the above works that aim to accelerate the SD models, some papers study the pruning methods on relatively small diffusion models. Diff-Pruning~\cite{fang2023structural}~prunes the channels of diffusion models by employing the dependency graph tool developed in DepGraph~\cite{fang2023depgraph} to address the structural dependency problem when pruning models with residual connections. The method is shown to be effective on small-scale datasets with a max image resolution of 256$\times$256, no SD models evaluated.

Our paper differs from the works above in that we focus on accelerating a practical \textbf{\textit{video}} diffusion model, while the works above mainly focus on \textbf{\textit{image}} diffusion models.

%% file: 03_method.tex
\section{Methodolgy}
\label{sec:methodology}
\subsection{Prelimiaries}
\label{sec:preliminaries}
% Introduce diffusion model, EDM
\textbf{Diffusion Models.}\label{sec:preliminaries:video_diffusion_models} 
A diffusion model (DM) consists of two major processes: 1. a forward diffusion process that adds noise to the input data iteratively and 2. a reverse denoising process that removes the predicted noise to recover the original input. In the continuous-time framework~\cite{song2021scorebased,karras2022elucidating}, let $p(\boldsymbol{x};\sigma)$ as the distribution obtained by adding Gaussian noise $\sigma$ to the data distribution $p_{data}(\boldsymbol{x}_0)$. With a large $\sigma_{max}$, the distribution $p(\boldsymbol{x};\sigma_{max})$ is close to the pure noise distribution. Then, the probability flow ordinary differential equation (PF-ODE) is defined as:
\begin{equation}\label{eq:pf-ode}
  \mathrm{d}\boldsymbol{x} = -\dot{\sigma}_t \, \sigma_t \, \nabla_{\boldsymbol{x}} \log p(\boldsymbol{x}, \sigma_t) \, \mathrm{d}t,
\end{equation}
where the score function $\nabla_{\boldsymbol{x}} \log p(\boldsymbol{x}, \sigma_t)$ is generally approximated by $\frac{D_\theta(\boldsymbol{x}; \sigma_t) - \boldsymbol{x}}{\sigma_t^2}$. In the EDM framework~\cite{karras2022elucidating,karras2024analyzing}, $D_\theta(\boldsymbol{x}_t,\sigma_t)$ is parameterized as follows:
\begin{equation}\label{eq:edm}
  D_{\theta}=c_{0}(t)\boldsymbol{x}_t+c_{1}(t)f_{\theta}(c_{2}(t)\boldsymbol{x}_t, c_{3}(t)),
\end{equation}
where $f_{\theta}$ is the learnable neural network trained by minimizing denoising error, and $c_{0}$, $c_{1}$, $c_{2}$, and $c_{3}$ are the time-dependent conditioning coefficients. The I2V models SF-V~\cite{zhangSFVSingleForward2024} and SVD~\cite{blattmannStableVideoDiffusion2023} are developed based on the EDM framework.

\noindent\textbf{Consistency Models.}\label{sec:preliminaries:consistency_models} Consistency models (CM)~\cite{songConsistencyModels2023,song2024improved} offers a new family of generative models which enforces the learned neural network $f_\theta(\boldsymbol{x}_t, t)$ maps the arbitrary noise input $\boldsymbol{x}_t$ to the clean data $\boldsymbol{x}_0$, and the CM follows the self-consistency property $f_\theta(\boldsymbol{x}_t, t) = f_\theta(\boldsymbol{x}_t^\prime, t^\prime)$, where $\boldsymbol{x}_t^\prime$ and $\boldsymbol{x}_t$ represent the samples on the ODE trajectory at different times $t$ and $t^\prime$. Following the preconditioned EDM framework, the CM sets the boundary condition as $c_{0}(0)=1$ and $c_{1}(0)=0$.

One way to train the CM is to use consistency distillation, which distills the pre-trained DMs by minimizing the consistency loss:
\begin{equation}
  \begin{aligned}
    \mathcal{L}_{CM}( \theta, \theta^- ; \Psi)= \mathbb{E} \left[ d\left( f_\theta \left( \boldsymbol{x}_{t_{n+k}}, t_{n+k},  \right), f_{\theta^-} \left( \hat{\boldsymbol{x}}_{t_n}^{\Psi, \omega}, t_n,\right) \right) \right],
  \end{aligned}
\end{equation}
where $d$ is the distance function, ${\boldsymbol{x}}_{t_n}^{\Psi, \omega}$ is the reversed data by ODE solver $\Psi$ with classifier-free guidance (CFG)~\cite{ho2021classifier} weight $\omega$, $n$ is the time step for pretrained DM, and $k$ is step interval. The T2V model T2V-Turbo-v2~\cite{song2024improved} is developed based on the CM.

\input{figs/fvd_time_param_i2v.tex}
\input{figs/framework.tex}
\subsection{Network Pruning}
\label{sec:method:network pruning}

\noindent\textbf{Individual Content and Motion Dynamics Preserved Pruning.}
Since the different network blocks play distinct roles in generating videos, we should assess the importance of blocks in both \emph{individual content quality} and \emph{motion dynamics}, followed by using the FVD score to quantify the block importance of the SF-V model.

By deconstructing the U-Net architecture, we assess the importance of each block by systematically replacing it with either an identity mapping or a single convolutional layer in cases where there is a channel mismatch. This approach allows us to isolate and evaluate the contribution of individual blocks
As illustrated in~\ref{fig:fvd_time_param_i2v}, ``D'', ``M'', and "U" denote the DownBlock, MidBlock, and UpBlock, respectively. For a specific block ``D.0.R.1'', ``R'' and ``A'' represent ResBlock and AttentionBlock, respectively. The first number ``0'' and the second number ``1'' represent the block index and the layer index within the block, respectively. 
Our overall evaluation indicate that removing low-resolution layers in the U-Net significantly raises the FVD score, suggesting a substantial impact on the generation quality of individual frames. In contrast, removing high-resolution layers has a comparatively smaller effect on the FVD score, indicating that these layers are less critical to maintaining the generation quality of each individual frame.
In addition to evaluating the individual frame generation quality, we also analyze the motion dynamics of the generated videos by visualizing the generated videos of the pruned U-Net. Our empirical observations reveal that removing the blocks in ``D.2'' and ``U.1'' significantly affects the temporal coherence and motion dynamics across frames, leading to noticeable degradation in video quality.

In the end, we prune the second R-A pairs in the DownBlocks and UpBlocks except for the second last DownBlock ``D.2'' and the second UpBlock ``U.1'', and further remove the MidBlock.  Further visualization results and detailed analyses are provided in the supplement.

\noindent\textbf{Compressed Network Architecture.} In addition to the U-Net, we find that the VAE decoder used in SF-V significantly contributes to the inference time during the latent decoding process. To address this, we apply both layer pruning and channel pruning to the VAE decoder. Utilizing the introduced network pruning techniques, we develop lightweight VDMini models for I2V and T2V tasks, referred to as VDMini-I2V and VDMini-T2V. Finally, the pruned models demonstrate a 2.5$\times$ speedup for I2V model SF-V and a 1.4$\times$ speedup for T2V model T2V-Turbo-v2. We extend our compression technique and achieve a 1.25$\times$ speedup on the DiT-based T2V model, HunyuanVideo.

\subsection{ICMD Consistency Loss}
\label{sec:method:ICMD loss}

We aim to further enhance the generation performance of the pruned VDMini network by mimicking the intermediate outputs of the U-Net from the original VDM. As previously mentioned, in order to maintain the generation quality of individual frames and the motion dynamics across the video,
we propose the Individual Content and Motion Dynamic (ICMD) Consistency Loss. This optimization strategy consists of two parts: Individual Content Distillation (ICD) Loss and Multi-frame Content Adversarial (MCA) Loss. The proposed framework is illustrated in Figure~\ref{fig:framework}.

\noindent\textbf{Individual Content Distillation Loss:}
As illustrated in~\ref{fig:framework}, ICD loss aims to achieve feature distillation~\cite{kimBKSDMLightweightFast2023} for each individual frame, which transfers the knowledge from the UNet of VDM (i.e., teacher model) to VDMini (i.e., student model). Formally, given a noisy video latent $\boldsymbol{x}_t \in R^{F \times C \times H \times W}$, the student model $f^{stu}$ is trained by minimizing the distance of intermediate representations between the teacher model $f^{tea}$ and the student model:
\begin{equation}
  \mathcal{L}_{ICD} = \mathbb{E} \left[ \sum_{l=1}^L d \left( f^{stu}_l \left( \boldsymbol{x}_t, t \right), f^{tea}_l \left( \boldsymbol{x}_t, t \right) \right) \right],
\end{equation}
where $L$ is the number of intermediate representation, and  $f^{stu}_l(\cdot)$ and $f^{tea}_l(\cdot)$ denote the feature representations at the $l$-th block of the student and teacher models, respectively.

\noindent\textbf{Multi-frame Content Adversarial Loss:}
With the ICD loss, the student model is forced to mimic the teacher model at the individual content level. However, the motion dynamics across the generated video can not be guaranteed. To address this issue, we introduce the MCA loss to adopt adversarial optimization strategy for preserving the motion dynamics across the generated video. Here, we let the discriminator $D_{\phi}$ to distinguish the video latent generated by the student model $f^{stu}$ from the output of the teacher model $f^{tea}$ as in Diffusion-GAN~\cite{wang2023diffusion}. The discriminator $D_{\phi}$ comprises a combination of SpatioHead and TemporalHead components with standard 1D and 2D convolutional operations shown in Figure~\ref{fig:framework}. We empirically observe that the temporal heads in the discriminator effectively enhance motion dynamics. Inspired by MCM and SF-V, we observe that the learnable MLP better captures motion in video generation tasks. Therefore, we incorporate these simple temporal heads to encode the motion information.

The optimization objective $\mathcal{L}_{MCA}$ can be formulated as:
\begin{equation}
  \begin{aligned}
    \mathcal{L}^{gen}_{MCA} &= -\mathbb{E} \left[ \log \left(D_{\phi} \left( f^{stu} \left( \boldsymbol{x}_t, t \right) \right), \sigma_{t^\prime} \right) \right], \\
    \mathcal{L}^{disc}_{MCA} &= \mathbb{E} \left[ \max(0, 1 + D_{\phi}(f^{stu}(\boldsymbol{x}_t), t), \sigma_{t^{\prime}}) \right] \\ % + \gamma R1 \right] \\
    &+ \mathbb{E} \left[ \max(0, 1 - D_{\phi}(f^{tea}(\boldsymbol{x}_t, t), \sigma_{t^{\prime}})) \right],
  \end{aligned}
\end{equation}
where $\sigma_{t^\prime}$ is the instance noise injected into the samples at timestep $t^\prime$, we follow SF-V to set the noise distribution as a discretized lognormal distribution, and $t^\prime \in [1, 999]$.
During SF-V fine-tuning, we incorporate both the teacher model and the original discriminator into the discriminator loss. This approach leverages real and synthetic samples from the teacher model to guide the score estimation of the denoiser, resulting in better performance in our experiments.

Along with our newly proposed ICMD loss, task-specific losses $\mathcal{L}_{Task}$ are also incorporated during the fine-tuning process.
Specifically, for the I2V task, SF-V employs a combination of reconstruction loss and adversarial loss. 
For the T2V task, T2V-Turbo-v2 utilizes a consistency loss along with a mixture of reward optimization objectives, 
and we further extend the proposed pruning and distillation method on the DiT-based T2V model HunyuanVideo~\cite{kong2024hunyuanvideo}, which trains a 13B model with the flow matching framework~\cite{lipman23flow}.

The final optimization objective is formulated as:
\begin{equation}
  \begin{aligned}
    \mathcal{L} = \mathcal{L}_{Task} + \lambda_{ICD} \mathcal{L}_{ICD} + \lambda_{MCA} (\mathcal{L}^{gen}_{MCA} + \mathcal{L}^{disc}_{MCA}),
  \end{aligned}
\end{equation}
where $\lambda_{ICD}$ and $\lambda_{MCA}$ are the hyper-parameters to balance the ICD and MCA loss, respectively, the optimization details about task-specific losses for the I2V and T2V tasks are provided in the supplementary materials. During SF-V fine-tuning, we incorporate both the original task-specific loss $\mathcal{L}_{Task}$ and the added $\mathcal{L}_{ICD}$ and $\mathcal{L}_{MCA}$ losses, the teacher model and the original discriminator are both utilized in the discriminator loss. This leverages real and synthetic samples from the teacher model to guide the score estimation of the denoiser, resulting in better performance in our experiments.

%% file: figs/fvd_time_param_i2v.tex
% Use figure* for multi-column figure

\begin{figure}[htbp]
    \vspace{-1.5em}
    \resizebox{\columnwidth}{!}{
        \centering
        \subfloat[]{
            \includegraphics[width=\linewidth]{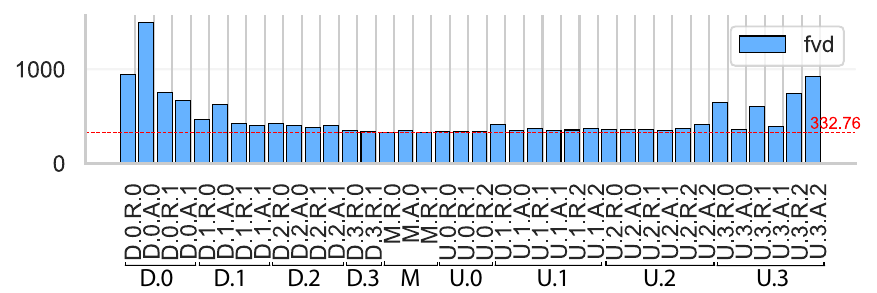}
        }
    }
    \resizebox{\columnwidth}{!}{
        \centering
        \subfloat[]{
            \includegraphics[width=\linewidth]{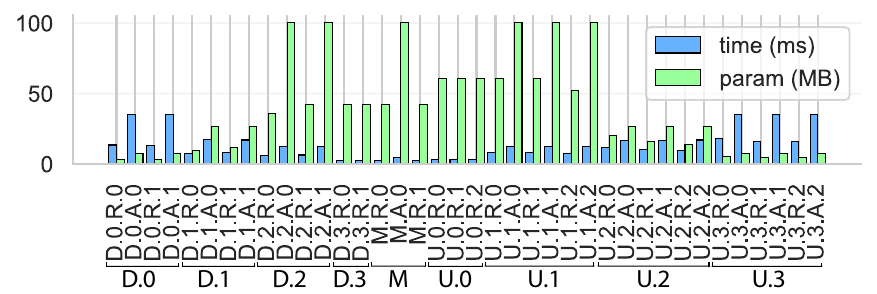}
        }
    }
    \vspace{-1.5em}
    \caption{(a) FVD score by removing or replacing the blocks in the U-Net. (\textbf{Note that a high FVD score means the block is more important.}) (b) Time and Parameters of the blocks in the U-Net.}
    \Description{FVD score for network pruning.}
    \vspace{-1em}
\label{fig:fvd_time_param_i2v}
\end{figure}

%% file: figs/framework.tex
% Use figure* for multi-column figure
\begin{figure*}[htbp]
    \centering
    \includegraphics[width=0.95\linewidth]{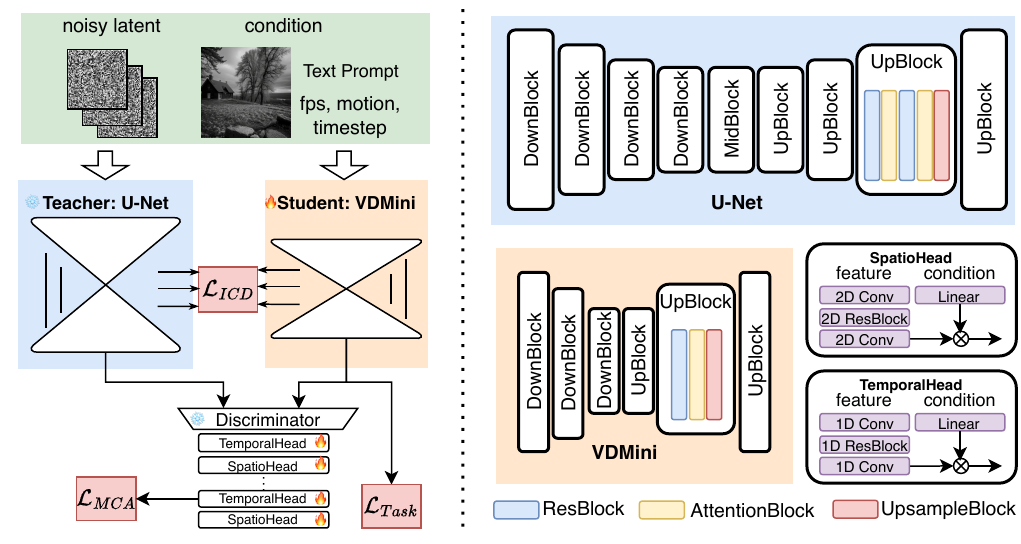}
    \vspace{-1em}
    \caption{The proposed VDMini framework for Video Diffusion Model Compression. Left: The retraining process with the proposed ICMD loss, where $\mathcal{L}_{ICD}$ is the knowledge distillation loss for individual content consistency, and $\mathcal{L}_{MCA}$ is the adversarial loss for multi-frame content consistency. $\mathcal{L}_{Task}$ is the task-specific loss function adopted in the base model. Right: The teacher model is pruned by blocks to obtain the student model (\ie VDMini). The second Block (ResBlock, AttentionBlock) in the DownBlock and UpBlock are removed (Except for the second last DownBlock and UpBlock), and the innermost Blocks (MidBlock, DownBlock, and UpBlock) are entirely removed.}% The student model is fine-tuned using the Individual and Integrated Content Consistency (ICMD) Loss, which consists of Individual Content Distillation (ICD) Loss and Integrated Content Adversarial (ICA) Loss.}
    \label{fig:framework}
    \vspace{-1em}
    \Description{VDMini framework}
\end{figure*}

%% file: 04_experiments.tex
\section{Experiments}
\label{sec:experiments}
\subsection{Datasets and Evaluation Metrics}
\label{sec:experiments:datasets_and_evaluation_metrics}
\noindent\textbf{Datasets:} We utilize three datasets for training: OpenVid-1M~\cite{nanOpenVid1MLargeScaleHighQuality2024}, VidGen-1M~\cite{tanVidGen1MLargeScaleDataset2024}, WebVid-10M~\cite {bain2021frozen} and HD-Mixkit~\cite{lin2024open}. OpenVid-1M and VidGen-1M are large-scale video datasets featuring high-quality videos and expressive captions, each containing over 1 million video-caption pairs. WebVid-10M is a widely-used video dataset collected from the web, encompassing diverse video content. For the I2V task, we first re-implement SF-V using the OpenVid-1M dataset, and then fine-tune the pruned VDMini-I2V model. For the T2V task, we follow the T2V-Turbo-v2~\cite{liT2VTurbov2EnhancingVideo2024} setup, training the model on a dataset of VidGen-1M and WebVid-10M, comprising 180K video-text pairs. 
For fine-tuning the HunyuanVideo model, we adopt the implementation from FastVideo\footnote{https://github.com/hao-ai-lab/FastVideo} and utilize a dataset comprising 1.2M samples from OpenVid-1M and HD-Mixkit.

\noindent\textbf{Evaluation Metrics:} We adopt the same evaluation metrics used in SF-V and T2V-Turbo-v2 for a fair comparison. Specifically, we use the FVD score~\cite{geContentBiasFrechet2024} to measure the quality of the generated videos in the I2V task, which calculates the Fréchet Video Distance between the generated videos and the ground-truth videos from UCF101. Following SF-V, we use the first frame of the video as the conditional input to VDMini to generate the video with a resolution of $768 \times 1024$, then resize the video to $240 \times 320$ for evaluation. The videos are saved as 95\% quality JPEG images, following the same setting in~\cite{skorokhodov2022stylegan}. And we evalute the I2V models on the comprehensive benchmark VBench-I2V~\cite{huang2024vbench++} to have a multi-dimensional evaluation. 
For the T2V task, we assess the performance of our model on VBench-T2V~\cite{huang2023vbench}, which automatically evaluates T2V models in terms of video quality and video prompt consistency, with a total of 16 diverse evaluation dimensions. We set the inference step to 16 for the T2V task, generating videos with a resolution of $320 \times 512$ and a video length of 16 for T2V-Turbo-v2, and the resolution for HunyuanVideo is 720p with 125 frames.

\input{tables/table_i2v_motion_analysis.tex}
\input{tables/table_i2v_comparison.tex}
\subsection{Implementation Details}
\label{sec:experiments:implementation_details}
In the experiments, we use 8 NVIDIA A100 GPUs for training the VDMini models and evaluate the models on a single A100 GPU. During the pruning stage, we evaluate the per-block FVD by replacing each block with either an identity block or a shortcut convolution layer regarding channel mismatch. We sample 1200 videos for evaluting in this stage, which takes around 9.5 hours on the server. 
For training the VDMini-I2V model, we set the resolution of the input video to $448 \times 576$. The model is trained with a batch size of 32 (with gradient accumulation of 4) for 10K steps, and the MCA loss is enabled after 3K steps. The learning rate for the U-Net and the discriminator is set to 1e-4 and 1e-5, respectively. The VAE decoder is trained with a batch size of 8 for 15K steps. The frame length is sampled from 14 to 25, and the learning rate is initially set to 5e-5 and decayed using a cosine annealing schedule. The loss weights for reconstruction loss, perceptual loss, and adversarial loss are set to 1, 1, and 0.2, respectively.
For training the VDMini-T2V model based on T2V-Turbo-v2, the resolution of the input video is set to $320 \times 512$. We set the batch size for fine-tuning to 2 with 8K steps of gradient updating, and the learning rate for the discriminator is set to 1e-5. All other hyper-parameters are kept the same as in the T2V-Turbo-v2 training. The loss weights for ICD loss and MCA loss are set to 0.1 and 1, respectively. More details about training VDMini-T2V model based on HunyuanVideo are presented in the supplement.

\input{tables/table_i2v_ablation_pruning.tex}
\input{figs/vdmini_i2v_animate.tex}
\input{figs/vdmini_t2v_turbo_animate.tex}
\subsection{Quantitative Results}
\label{sec:experiments:comparison with sota methods}

\noindent\textbf{Comparison with Other I2V Methods.}
To evaluate the performance of VDMini-I2V, we compare it with several state-of-the-art I2V models, including SF-V~\cite{skorokhodov2022stylegan}, SVD~\cite{blattmannStableVideoDiffusion2023}, and AnimateLCM~\cite{wangAnimateLCMAcceleratingAnimation2024}. The results are reported in Table~\ref{tab:i2v_comparison} and Table~\ref{tab:tab_i2v_motion_analysis}.

As reported in Table~\ref{tab:i2v_comparison}, VDMini-I2V compresses the U-Net by approximately 40\% in terms of parameters and achieves an FVD score of 198.13 after retraining, with no significant drop compared to the teacher model (\ie SF-V). When compared to SVD with 16 steps, VDMini-I2V achieves a similar FVD score but is much faster, with a 37$\times$ speedup. This substantial speedup is due to the pruning and fine-tuning strategy employed by VDMini, which reduces the computational complexity of the model without compromising the quality of the generated videos. The reduction in parameters not only accelerates inference time but also decreases the memory footprint, making VDMini-I2V more suitable for deployment in resource-constrained environments. Additionally, the retraining process ensures that the pruned model regains high fidelity in video generation, maintaining the visual quality and temporal coherence of the output videos. This balance between efficiency and performance highlights the practical advantages of VDMini-I2V over other SoTA methods.

Except for the evaluation on the UCF101 dataset, we further evaluate the performance of VDMini-I2V on the VBench-I2V benchmark~\cite{huang2024vbench++}, which provides a comprehensive evaluation of I2V models across multiple dimensions. 
As reported in Table~\ref{tab:tab_i2v_motion_analysis}, we observe that VDMini-I2V achieves competitive performance across various subjects, backgrounds, and motion-related metrics. The results indicate that VDMini-I2V effectively preserves the quality of the generated videos while significantly reducing the model size and inference time. This demonstrates the effectiveness of our proposed method in enhancing the performance of I2V models.
The results of the motion-related metrics, \textbf{Motion Smoothness} and \textbf{Dynamic Degree}, in Table~\ref{tab:tab_i2v_motion_analysis} present that $\mathcal{L}_{MCA}$ effectively preserves motion dynamics during retraining. VDMini-I2V demonstrates competitive performance compared to the unpruned model SF-V, particularly excelling in motion-related metrics such as Motion Smoothness and Dynamic Degree.

\noindent\textbf{Comparison with Other Baseline Pruning Methods.} 
To further validate the effectiveness of the proposed VDMini method, we compare its performance with other structural pruning methods, including Magnitude-based pruning~\cite{li2017pruning} and DepGraph~\cite{fang2023depgraph}. Magnitude-based pruning focuses on removing the smallest magnitude weights in the network, while DepGraph constructs a dependency graph for the network and iteratively prunes model weights with low-importance scores, using the L2 Norm and first-order Taylor expansion of loss as the importance score. As shown in Table~\ref{tab:i2v_ablation_pruning}, our block pruning approach achieves faster latency than these methods for the same model size, primarily because most inference time is spent on high-resolution layers, such as the first DownBlock and the last UpBlock. Since the importance scores used by these pruning methods are not directly correlated with the evaluation protocol, the pruned network architecture may be suboptimal. With the same retraining process, VDMini-I2V achieves a better FVD score compared to these pruning methods.

\input{tables/table_i2v_ablation.tex}
\noindent\textbf{Effectiveness of the ICMD Loss.}
\label{sec:experiments:effectiveness_of_ICMD_loss}
We conduct ablation studies to validate the effectiveness of the ICMD loss used during the fine-tuning stage. As shown in Table~\ref{tab:i2v_ablation_loss}, enabling $\mathcal{L}_{ICD}$ and $\mathcal{L}_{MCA}$ individually results in FVD scores of 224.24 and 257.99, respectively. When both $\mathcal{L}_{ICD}$ and $\mathcal{L}_{MCA}$ are combined, the FVD score improves significantly to 198.13. These results highlight the substantial contribution of the ICMD loss to the overall performance of the model.

\nbf{Effectiveness of Loss Weight} To balance the ICMD loss with the task-specific loss during the training stage, we perform experiments with different hyper-parameters $\lambda_{ICD}$ and $\lambda_{MCA}$. As shown in Table~\ref{tab:i2v_ablation_lambda}, we find that the result is sensitive to $\lambda_{ICD}$, while more robust to $\lambda_{MCA}$. The best result is obtained when $\lambda_{ICD}$ and $\lambda_{MCA}$ are set to 0.1 and 1, respectively. This indicates that careful tuning of these hyper-parameters is crucial for optimizing the performance of the model.

\input{tables/table_t2v_comparison.tex}
\noindent\textbf{Comparison with Other T2V models.} On the other hand, we apply the proposed pruning and fine-tuning strategy on UNet-based T2V-Turbo-v2 and DiT-based HunyuanVideo, resulting in VDMini-T2V. As shown in  Table~\ref{tab:t2v_comparison}, we find that VDMini-T2V (T2V-Turbo-v2) reduces 35\% inference time compared to T2V-Turbo-v2, with a small degradation on Quality Score after retraining. However, the VDMini-T2V still achieves better results compared to the other models, such as VideoCrafter2 and KLing. Besides, we conduct experiments on DiT-based model HunyuanVideo by pruning the Dual-stream blocks from \textbf{20} to \textbf{12} and the Single-stream blocks from \textbf{40} to \textbf{28}, resulting in only a marginal performance drop of 0.82\% on VBench-T2V, while achieving a 1.25$\times$ speed up, the inference speed is tested on a single A100 GPU.

\subsection{Qualitative Results}
\label{sec:experiments:qualitative_results}
\nbf{Visual Results on the I2V Task} As shown in Figure~\ref{fig:i2v_animate}, the generated results of our method and other related approaches in the I2V task demonstrate satisfactory performance. The visual results indicate that our method effectively maintains both individual and multi-frame content consistency across frames, preserving similar styles and motions as the teacher model SF-V. In contrast, other methods suffer from blurring or distortion due to motion. Our method exhibits high visual quality and smooth motion with only one step inference.

\noindent\textbf{Visual Results on the T2V task.} We also present the visual results of the VDMini-T2V models in Figure~\ref{fig:t2v_turbo_animate}. The generated videos demonstrate that the VDMini-T2V model is capable of producing high-quality videos that accurately reflect a wide range of input text prompts. The visual consistency and fidelity of the generated videos highlight the effectiveness of our proposed method, ensuring that the VDMini-T2V model maintains the integrity and coherence of the video content while significantly reducing inference time.

%% file: tables/table_i2v_motion_analysis.tex
\begin{table*}[htbp]
  \centering
  \caption{Evaluation of VDMini-I2V on the VBench-I2V dataset. In this table, we compare the performance of the unpruned model SF-V and VDMini-I2V with and without the motion consistency loss $\mathcal{L}_{MCA}$. The metrics are divided into two categories: I2V subject and background consistency, and motion smoothness, dynamic degree, aesthetic quality, and imaging quality. The results show that VDMini-I2V achieves comparable performance to SF-V while being more efficient.}
  \label{tab:tab_i2v_motion_analysis}
  \vspace{-1em}
  \scalebox{0.99}{
    \begin{tabular}{l|c c c c}
      \toprule
      Models & \textbf{I2V Subject} $\uparrow$ & \textbf{I2V Background} $\uparrow$& \textbf{Subject Consistency} $\uparrow$& \textbf{Background Consistency} $\uparrow$ \\
      \midrule
      SF-V & 97.48\% & 97.59\% & 95.54\% & 96.63\% \\
      VDMini-I2V w/o $\mathcal{L}_{MCA}$ & 97.40\% & 97.36\% & 95.33\% & 96.12\% \\
      VDMini-I2V & 97.51\% & 97.53\% & 95.59\% & 96.54\% \\
      \midrule
      Models & \textbf{Motion Smoothness}$\uparrow$ & \textbf{Dynamic Degree}$\uparrow$ & \textbf{Aesthetic Quality}$\uparrow$ & \textbf{Imaging Quality}$\uparrow$ \\
      \midrule
      SF-V & 98.11\% & 32.13\% & 59.89\% & 68.48\% \\
      VDMini-I2V w/o $\mathcal{L}_{MCA}$ & 96.84\% & 23.84\% & 57.85\% & 66.25\% \\
      VDMini-I2V & 98.34\% & 33.10\% & 59.18\% & 67.99\% \\
      \bottomrule
    \end{tabular}
    }
  \vspace{-1em}
\end{table*}

%% file: tables/table_i2v_comparison.tex
\begin{table}[htbp]
  % \vspace{-1em}
  \centering
  \caption{Comparison with the existing methods. VDMini-I2V achieves a comparable FVD score with SF-V and 16-step SVD.}
  \vspace{-0.5em}
  \label{tab:i2v_comparison}
  \resizebox{\columnwidth}{!}{
    \begin{tabular}{l|cccc}
    \toprule
    \textbf{Methods} & \textbf{NFE} $\downarrow$ & \textbf{FVD} $\downarrow$ & \textbf{Model Size$\downarrow$} & \textbf{Latency (ms)$\downarrow$} \\
    \hline
    SVD                     & 25   & 163.43                    & 1.5B                & 20126 \\
    SVD                     & 16   & 183.98                    & 1.5B                & 12880 \\
    AnimateLCM              & 8    & 320.71                    & 1.5B                & 7346 \\
    SF-V                    & 1    & 166.26                    & 1.5B                & 512 \\
    \hline
    VDMini-I2V      &  1           & 198.13                    & 940M                & 345 \\
    \bottomrule
    \end{tabular}
  }
  \vspace{-0.5em}
\end{table}

%% file: tables/table_i2v_ablation_pruning.tex
\begin{table}[t]
  % \vspace{-1em}
  \centering
  \caption{
    Comparison with other baseline pruning approaches. DepGraph~\cite{fang2023depgraph} explicitly models the dependency between layers and comprehensively groups coupled parameters for pruning, while Magnitude Pruning aims to remove the smallest magnitude weights in the network. L2 and Taylor are the criteria for importance estimation.
  }
  \vspace{-0.5em}
  \label{tab:i2v_ablation_pruning}
  % \scalebox{0.99}{
    \begin{tabular}{l cccc}
    \toprule
    \textbf{Methods} & \textbf{FVD} $\downarrow$ & \textbf{Latency (ms)} $\downarrow$\\\hline
    DepGraph (L2)  & 288.79    & 433 \\
    DepGraph (Taylor) & 292.18    & 425 \\
    Magnitude Pruning (L2) & 271.55    & 412 \\
    Magnitude Pruning (Taylor) & 354.51    & 430 \\\hline
    VDMini-I2V      & \textbf{198.13}    & \textbf{345} \\
    \bottomrule
    \end{tabular}
  % }
  \vspace{-0.5em}
\end{table}

%% file: figs/vdmini_i2v_animate.tex
\begin{figure*}[ht]
    
  \renewcommand{\numColumns}{6}
  \renewcommand{\columnSpacing}{0.15em}
  \begin{tabular}{    @{} 
    p{\dimexpr(\textwidth-\columnSpacing*(\numColumns-1))/\numColumns} @{\hspace{\columnSpacing}}
    p{\dimexpr(\textwidth-\columnSpacing*(\numColumns-1))/\numColumns} @{\hspace{\columnSpacing}}
    p{\dimexpr(\textwidth-\columnSpacing*(\numColumns-1))/\numColumns} @{\hspace{\columnSpacing}}
    p{\dimexpr(\textwidth-\columnSpacing*(\numColumns-1))/\numColumns} @{\hspace{\columnSpacing}}
    p{\dimexpr(\textwidth-\columnSpacing*(\numColumns-1))/\numColumns} @{\hspace{\columnSpacing}}
    p{\dimexpr(\textwidth-\columnSpacing*(\numColumns-1))/\numColumns} @{}
    }
    % \centering {{\vspace{-1em}SVD}} &
        \includegraphics[width=\linewidth]{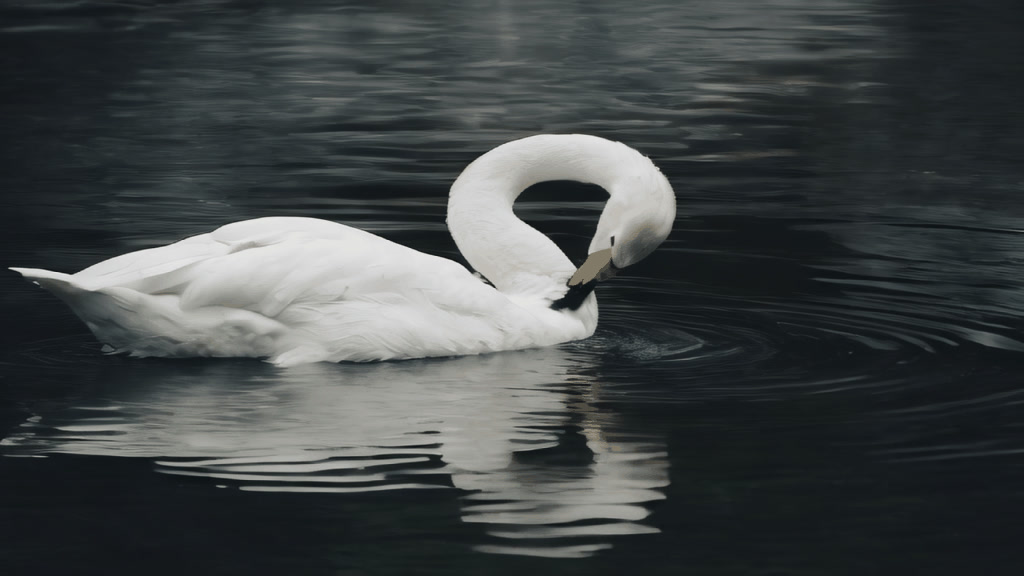} &
        \includegraphics[width=\linewidth]{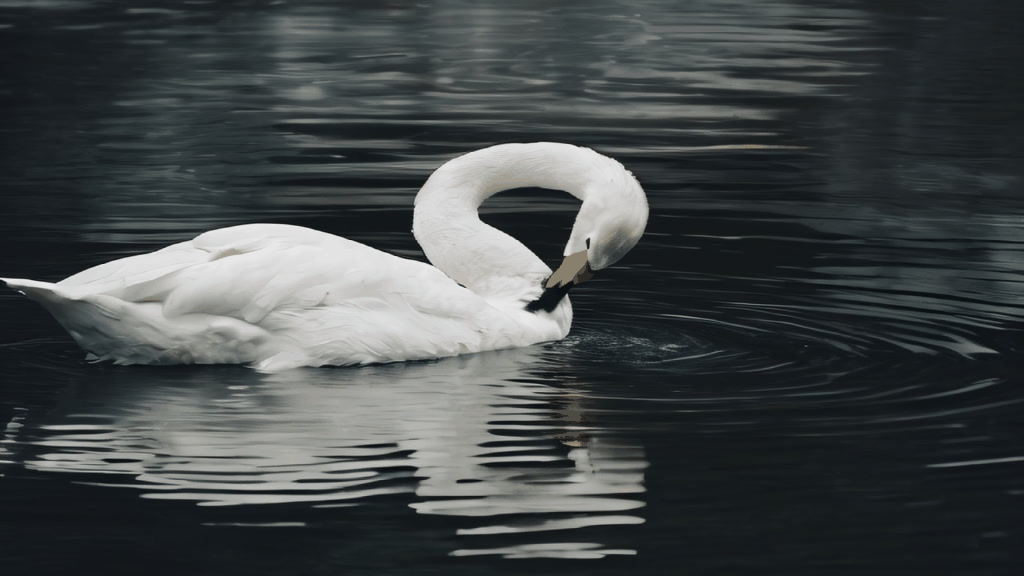} &
        {\animategraphics[width=\linewidth]{8}{figs/i2v_animates/svd/bird-7586857_1280_svd/}{01}{14}} &
        \includegraphics[width=\linewidth]{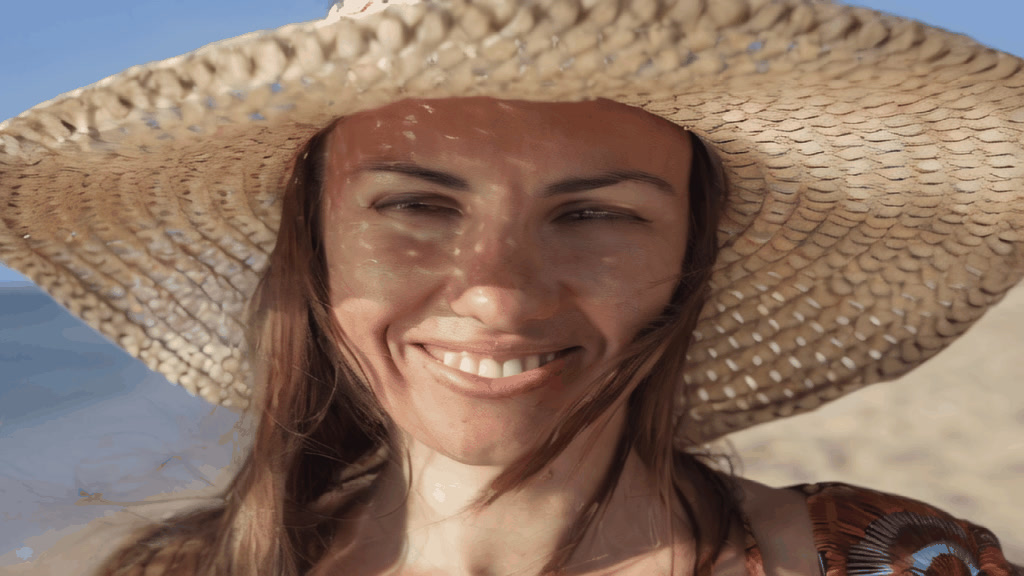} &
        \includegraphics[width=\linewidth]{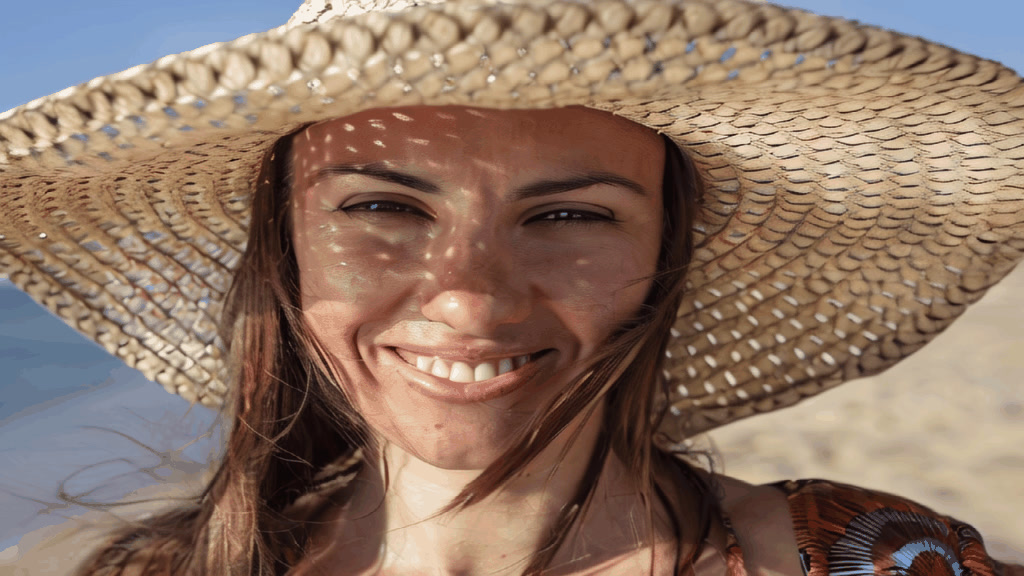} &
        {\animategraphics[width=\linewidth]{8}{figs/i2v_animates/svd/woman-4549327_1280_svd/}{01}{14}}
  \end{tabular}

  \begin{tabular}{
    @{}
    p{\dimexpr(\textwidth-\columnSpacing*(\numColumns-1))/\numColumns} @{\hspace{\columnSpacing}}
    p{\dimexpr(\textwidth-\columnSpacing*(\numColumns-1))/\numColumns} @{\hspace{\columnSpacing}}
    p{\dimexpr(\textwidth-\columnSpacing*(\numColumns-1))/\numColumns} @{\hspace{\columnSpacing}}
    p{\dimexpr(\textwidth-\columnSpacing*(\numColumns-1))/\numColumns} @{\hspace{\columnSpacing}}
    p{\dimexpr(\textwidth-\columnSpacing*(\numColumns-1))/\numColumns} @{\hspace{\columnSpacing}}
    p{\dimexpr(\textwidth-\columnSpacing*(\numColumns-1))/\numColumns} @{}
    }

        \includegraphics[width=\linewidth]{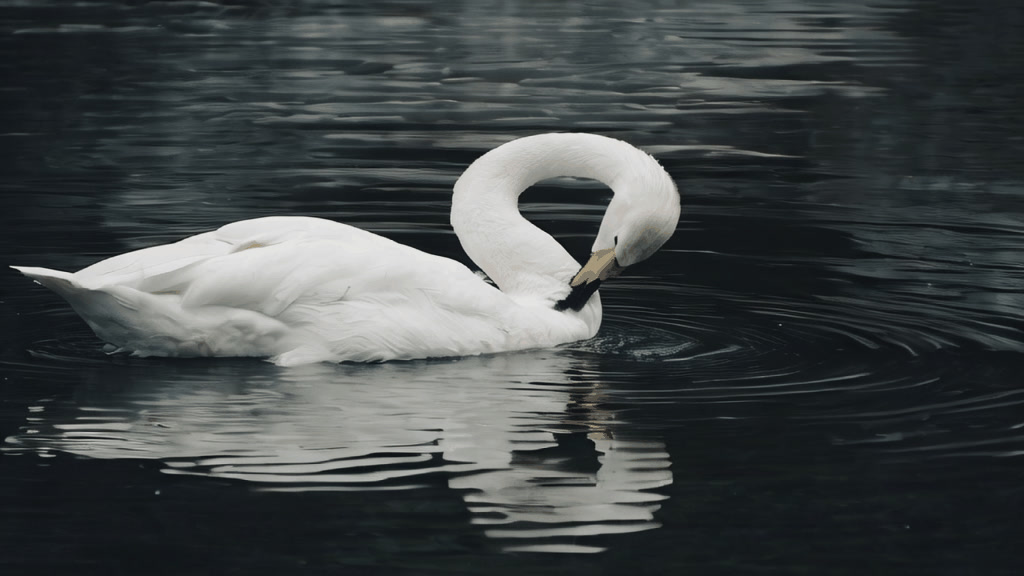} &
        \includegraphics[width=\linewidth]{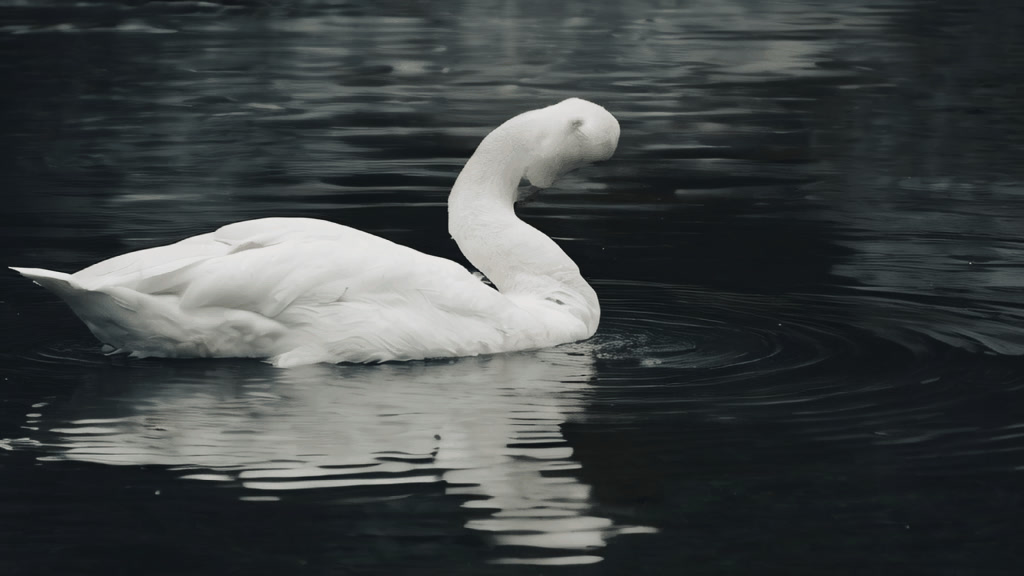} &
        {\animategraphics[width=\linewidth]{8}{figs/i2v_animates/animatelcm/bird-7586857_1280/}{01}{14}} &
        \includegraphics[width=\linewidth]{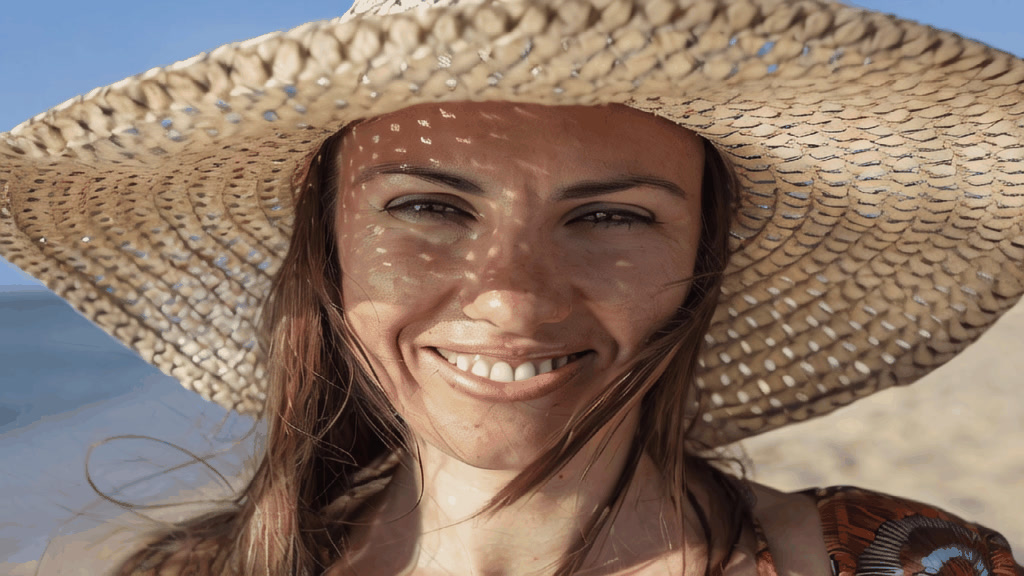} &
        \includegraphics[width=\linewidth]{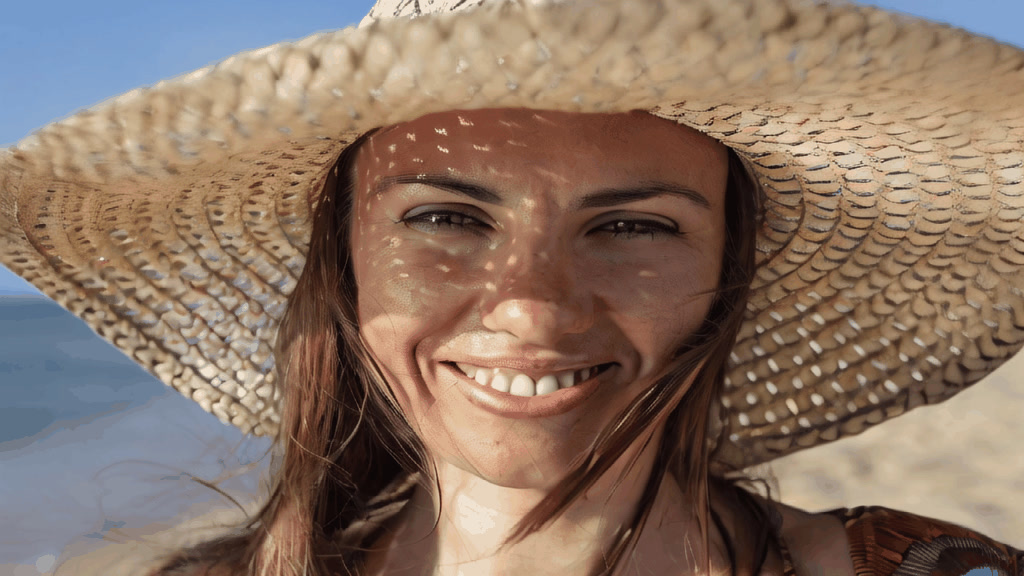} &
        {\animategraphics[width=\linewidth]{8}{figs/i2v_animates/animatelcm/woman-4549327_1280/}{01}{14}}
  \end{tabular}

  \begin{tabular}{
    @{}
    p{\dimexpr(\textwidth-\columnSpacing*(\numColumns-1))/\numColumns} @{\hspace{\columnSpacing}}
    p{\dimexpr(\textwidth-\columnSpacing*(\numColumns-1))/\numColumns} @{\hspace{\columnSpacing}}
    p{\dimexpr(\textwidth-\columnSpacing*(\numColumns-1))/\numColumns} @{\hspace{\columnSpacing}}
    p{\dimexpr(\textwidth-\columnSpacing*(\numColumns-1))/\numColumns} @{\hspace{\columnSpacing}}
    p{\dimexpr(\textwidth-\columnSpacing*(\numColumns-1))/\numColumns} @{\hspace{\columnSpacing}}
    p{\dimexpr(\textwidth-\columnSpacing*(\numColumns-1))/\numColumns} @{}
    }

        \includegraphics[width=\linewidth]{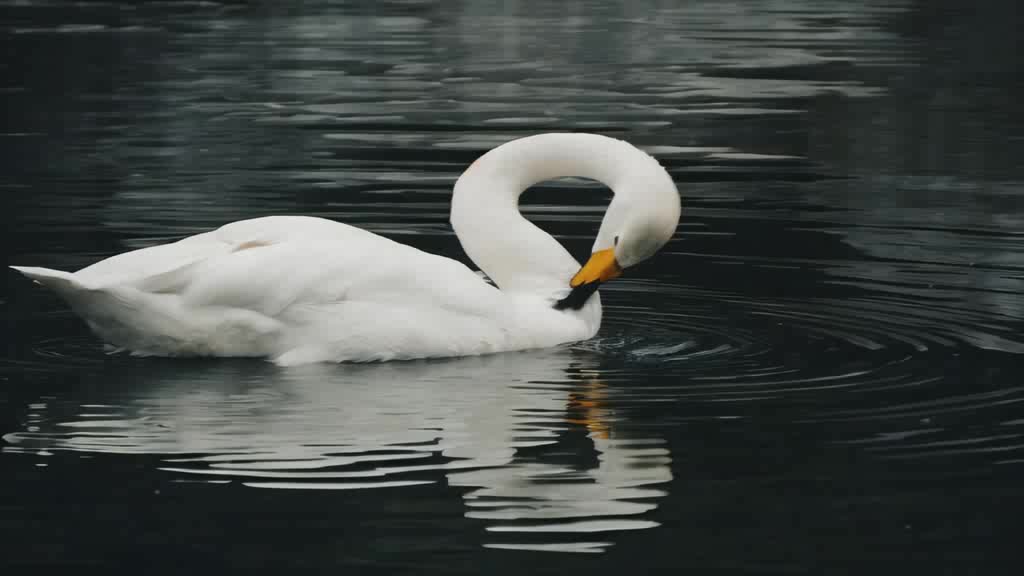} &
        \includegraphics[width=\linewidth]{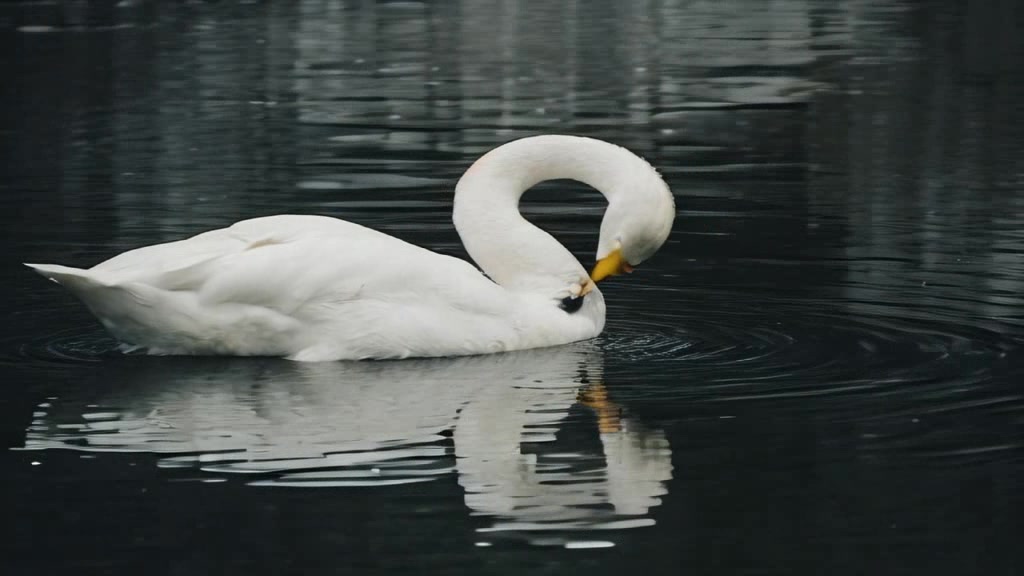} &
        { \animategraphics[width=\linewidth]{8}{figs/i2v_animates/sfv/bird-7586857_1280/}{01}{14}} &
        \includegraphics[width=\linewidth]{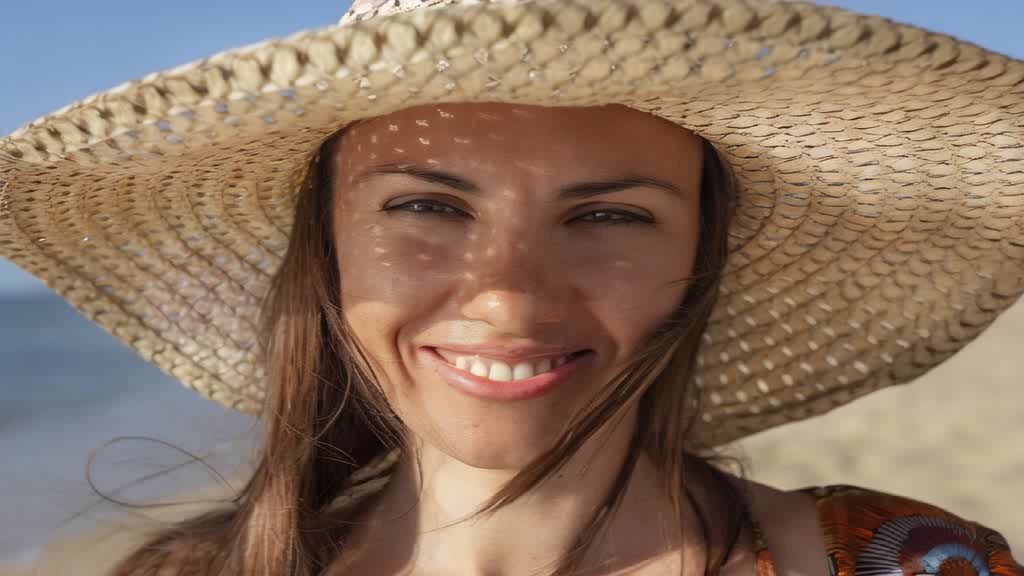} &
        \includegraphics[width=\linewidth]{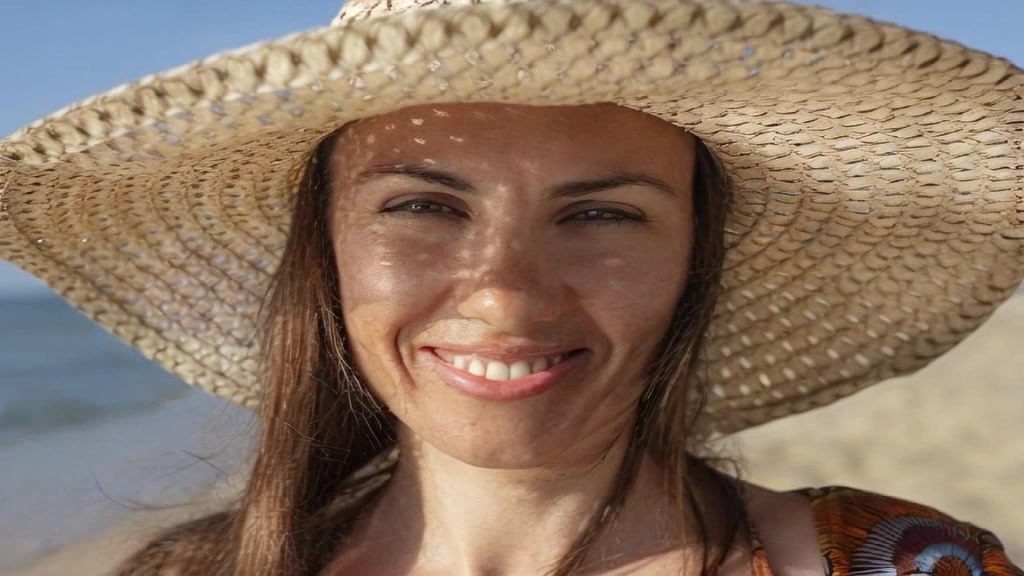} &
        {\animategraphics[width=\linewidth]{8}{figs/i2v_animates/sfv/woman-4549327_1280/}{01}{14}}
  \end{tabular}

  \begin{tabular}{
    @{}
    p{\dimexpr(\textwidth-\columnSpacing*(\numColumns-1))/\numColumns} @{\hspace{\columnSpacing}}
    p{\dimexpr(\textwidth-\columnSpacing*(\numColumns-1))/\numColumns} @{\hspace{\columnSpacing}}
    p{\dimexpr(\textwidth-\columnSpacing*(\numColumns-1))/\numColumns} @{\hspace{\columnSpacing}}
    p{\dimexpr(\textwidth-\columnSpacing*(\numColumns-1))/\numColumns} @{\hspace{\columnSpacing}}
    p{\dimexpr(\textwidth-\columnSpacing*(\numColumns-1))/\numColumns} @{\hspace{\columnSpacing}}
    p{\dimexpr(\textwidth-\columnSpacing*(\numColumns-1))/\numColumns} @{}
    }

        \includegraphics[width=\linewidth]{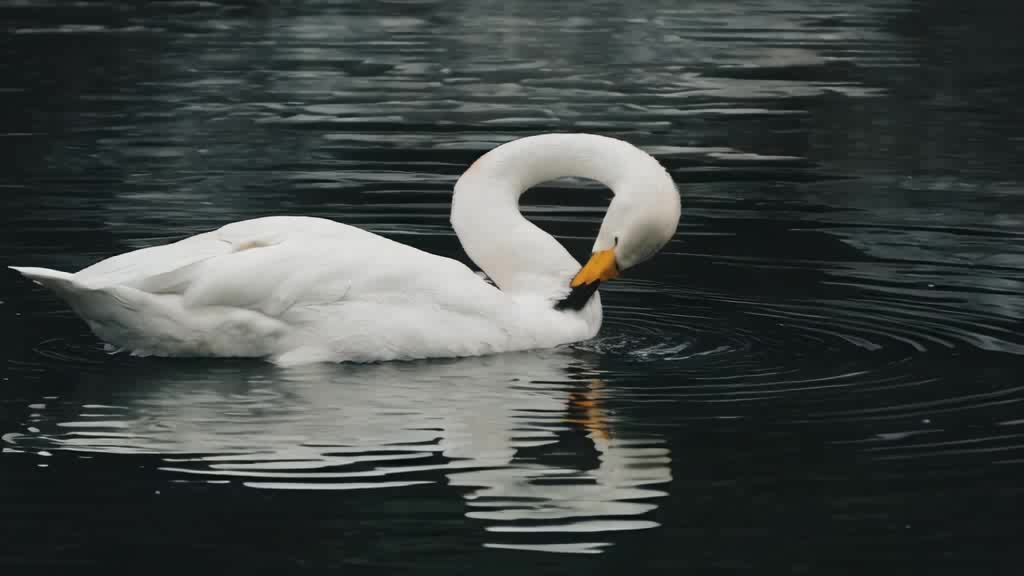} &
        \includegraphics[width=\linewidth]{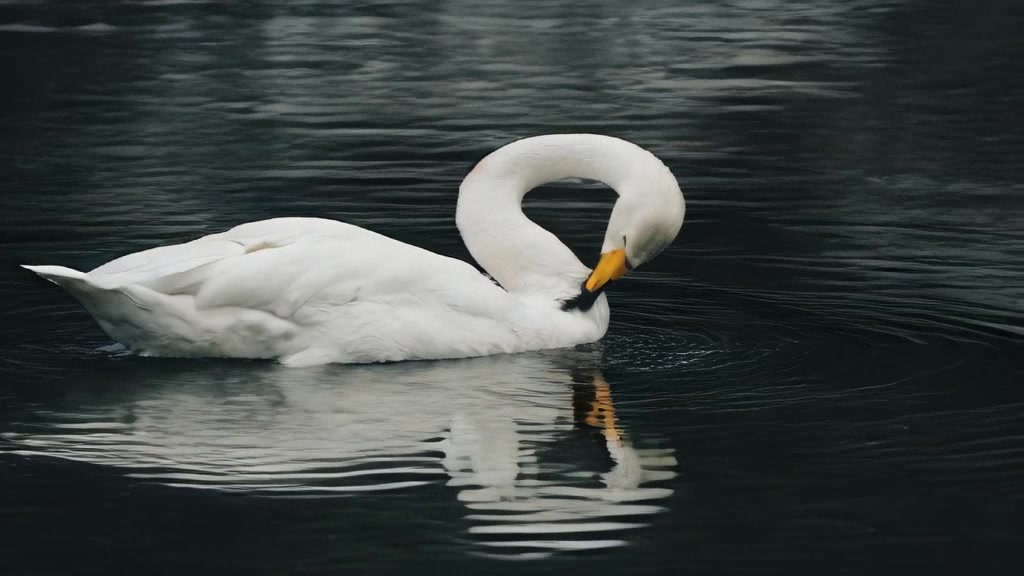} &
        {\animategraphics[width=\linewidth]{8}{figs/i2v_animates/vdmini/bird-7586857_1280/}{01}{14}} &
        \includegraphics[width=\linewidth]{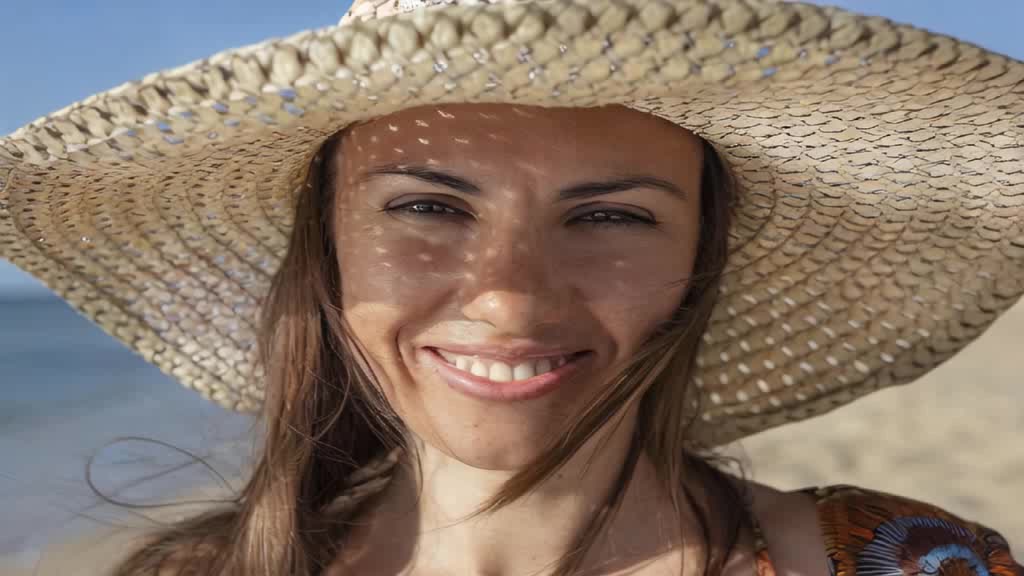} &
        \includegraphics[width=\linewidth]{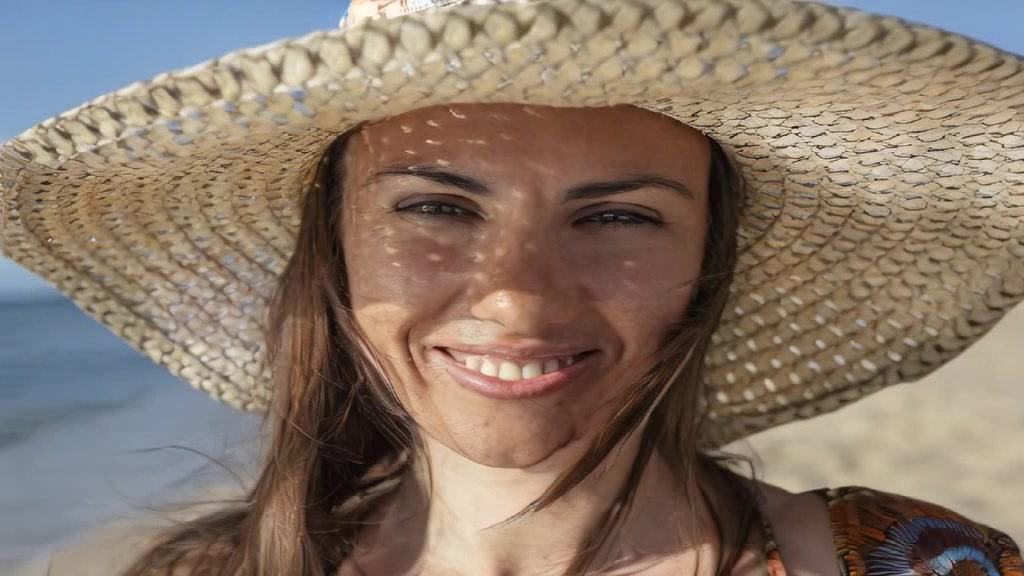} &
        {\animategraphics[width=\linewidth]{8}{figs/i2v_animates/vdmini/woman-4549327_1280/}{01}{14}}
  \end{tabular}
  \vspace{-1em}
  \caption{Qualitative results of VDMini-I2V based on SVD. From top to bottom, the rows correspond to videos generated by SVD (25 steps), AnimateLCM (8 steps), SF-V (1 step), and VDMini-I2V (1 step). For each video, the leftmost image is the prompt image. From left to right: the 1st frame, the 10th frame, and the video are shown. \red{\animationNotes}.}
  \Description{qualitative results of VDMini-I2V}
  \label{fig:i2v_animate}
  \vspace{-1em}
\end{figure*}

%% file: figs/vdmini_t2v_turbo_animate.tex
\begin{figure*}[ht]
  \begin{tabular}{
    @{} 
    p{\dimexpr(\textwidth-\columnSpacing*(\numColumns-1))/\numColumns} @{\hspace{\columnSpacing}}
    p{\dimexpr(\textwidth-\columnSpacing*(\numColumns-1))/\numColumns} @{\hspace{\columnSpacing}}
    p{\dimexpr(\textwidth-\columnSpacing*(\numColumns-1))/\numColumns} @{\hspace{\columnSpacing}}
    p{\dimexpr(\textwidth-\columnSpacing*(\numColumns-1))/\numColumns} @{}
  }
      \includegraphics[width=\linewidth]{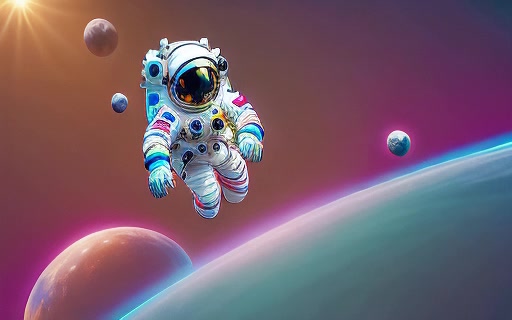} &
      \includegraphics[width=\linewidth]{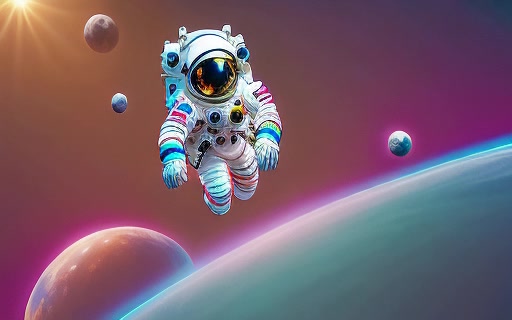} &
      \includegraphics[width=\linewidth]{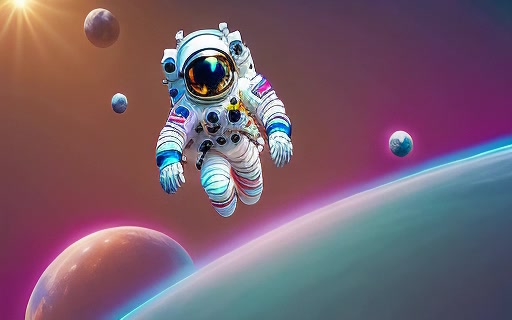} &
      {\animategraphics[width=\linewidth]{8}{figs/t2v_animates/t2v_turbo_v2/appearance_style/An_astronaut_flying_in_space,_surrealism_style/}{01}{16}}
  \end{tabular}

  \begin{tabular}{
    @{} 
    p{\dimexpr(\textwidth-\columnSpacing*(\numColumns-1))/\numColumns} @{\hspace{\columnSpacing}}
    p{\dimexpr(\textwidth-\columnSpacing*(\numColumns-1))/\numColumns} @{\hspace{\columnSpacing}}
    p{\dimexpr(\textwidth-\columnSpacing*(\numColumns-1))/\numColumns} @{\hspace{\columnSpacing}}
    p{\dimexpr(\textwidth-\columnSpacing*(\numColumns-1))/\numColumns} @{}
  }

      \includegraphics[width=\linewidth]{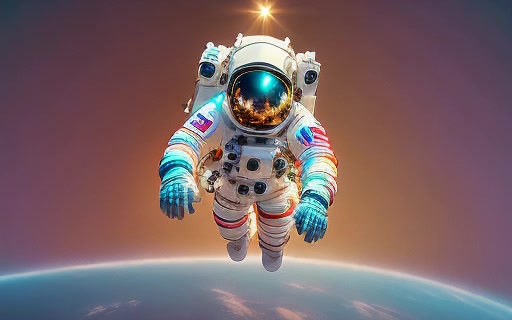} &
      \includegraphics[width=\linewidth]{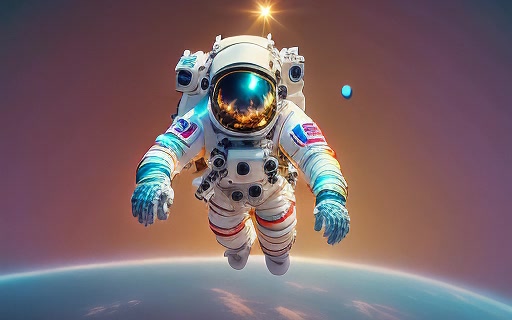} &
      \includegraphics[width=\linewidth]{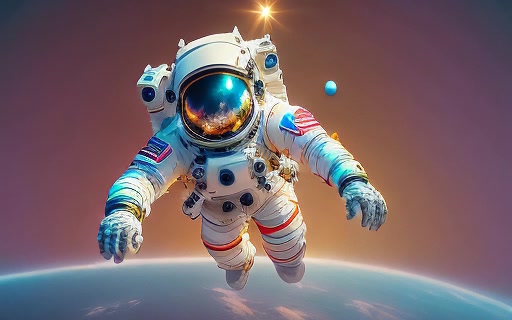} &
      {\animategraphics[width=\linewidth]{8}{figs/t2v_animates/vdmini/appearance_style/An_astronaut_flying_in_space,_surrealism_style/}{01}{16}}
  \end{tabular}
  \begin{tabular}{
    @{} 
    p{\dimexpr(\textwidth-\columnSpacing*(\numColumns-1))/\numColumns} @{\hspace{\columnSpacing}}
    p{\dimexpr(\textwidth-\columnSpacing*(\numColumns-1))/\numColumns} @{\hspace{\columnSpacing}}
    p{\dimexpr(\textwidth-\columnSpacing*(\numColumns-1))/\numColumns} @{\hspace{\columnSpacing}}
    p{\dimexpr(\textwidth-\columnSpacing*(\numColumns-1))/\numColumns} @{}
  }
      \includegraphics[width=\linewidth]{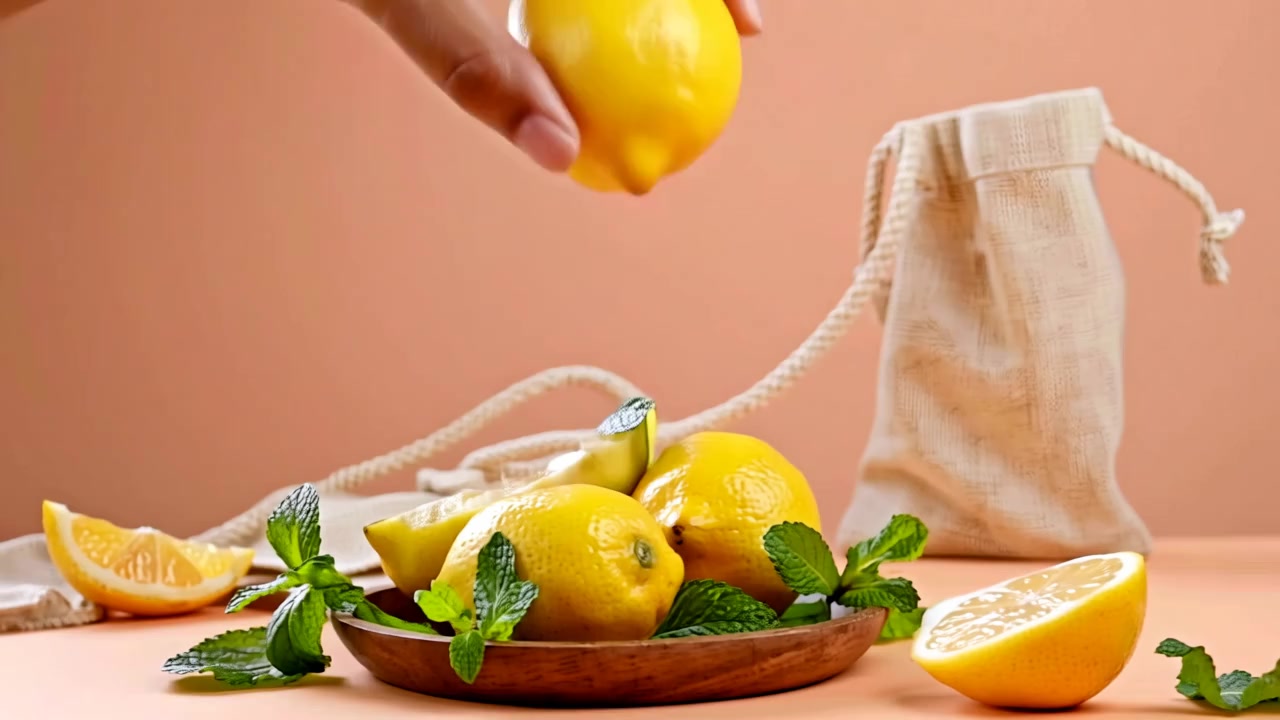} &
      \includegraphics[width=\linewidth]{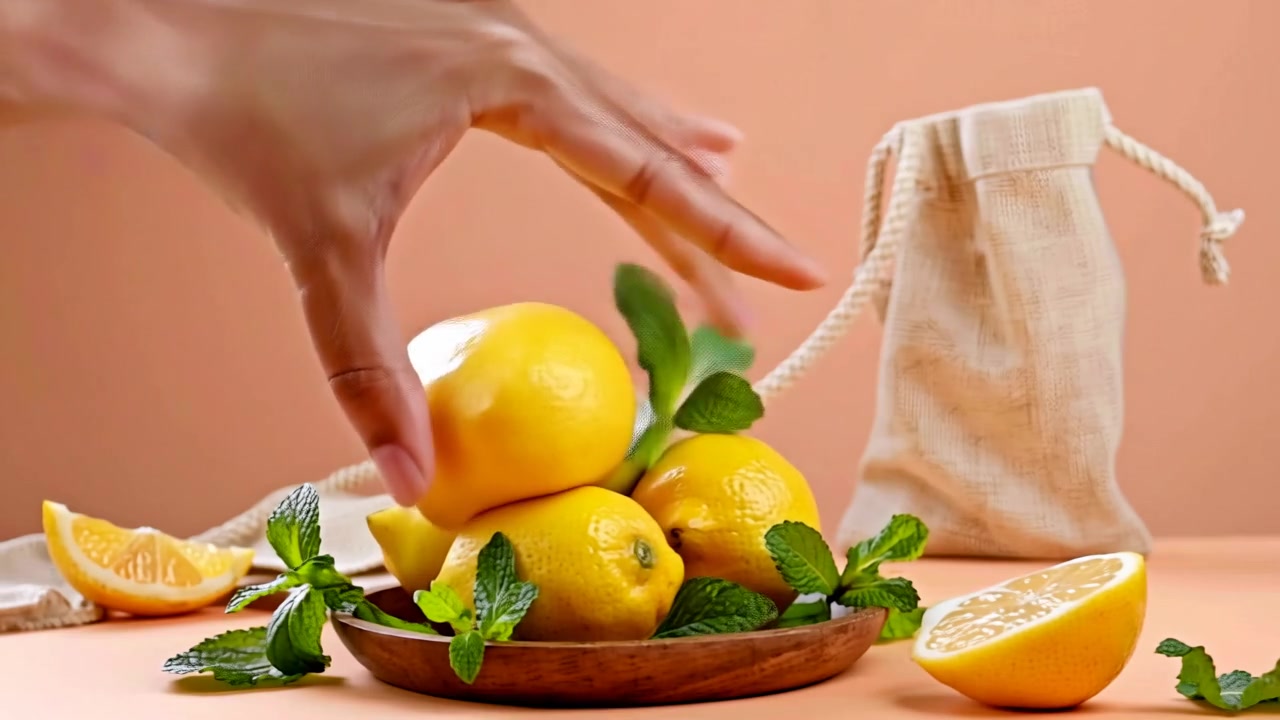} &
      \includegraphics[width=\linewidth]{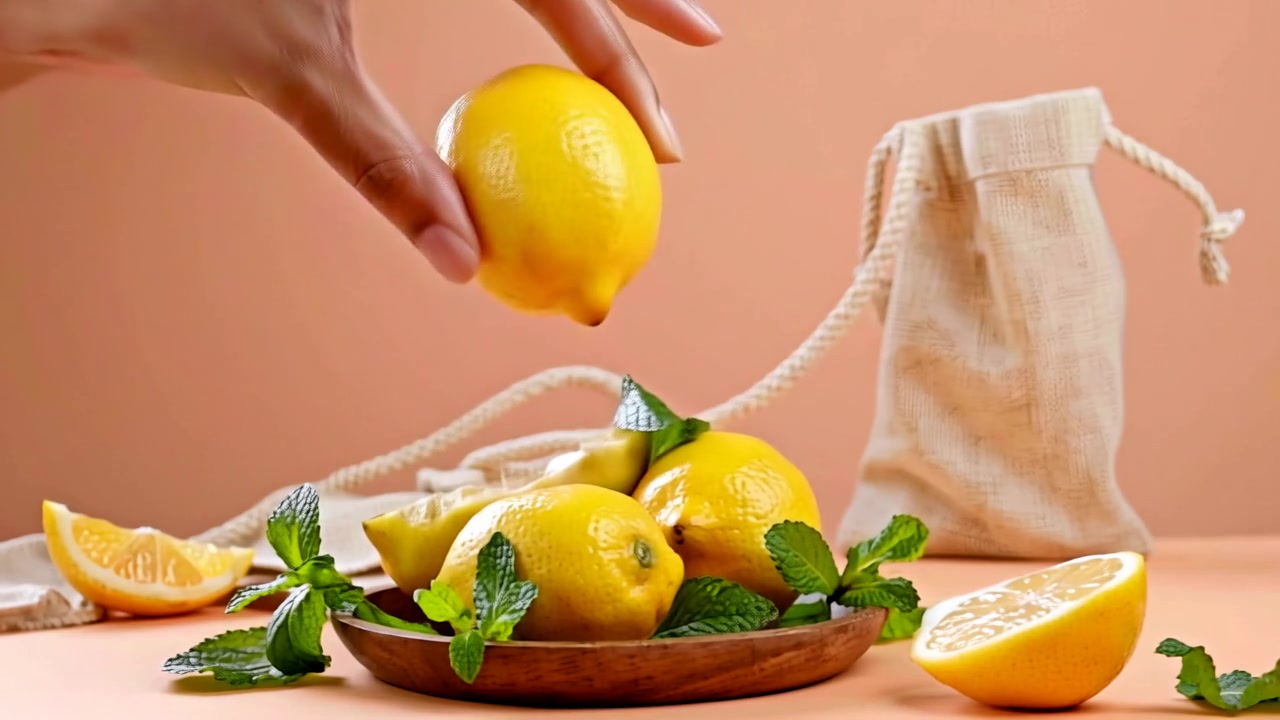} &
      {\animategraphics[width=\linewidth]{8}{figs/t2v_animates/hunyuanvideo/A_hand_with_delicate_fingers_picks_up_a_bright_yellow_lemon_from_a_wooden_bowl_filled_with_lemons_an/}{001}{125}}
  \end{tabular}

  \begin{tabular}{
    @{} 
    p{\dimexpr(\textwidth-\columnSpacing*(\numColumns-1))/\numColumns} @{\hspace{\columnSpacing}}
    p{\dimexpr(\textwidth-\columnSpacing*(\numColumns-1))/\numColumns} @{\hspace{\columnSpacing}}
    p{\dimexpr(\textwidth-\columnSpacing*(\numColumns-1))/\numColumns} @{\hspace{\columnSpacing}}
    p{\dimexpr(\textwidth-\columnSpacing*(\numColumns-1))/\numColumns} @{}
  }

        \includegraphics[width=\linewidth]{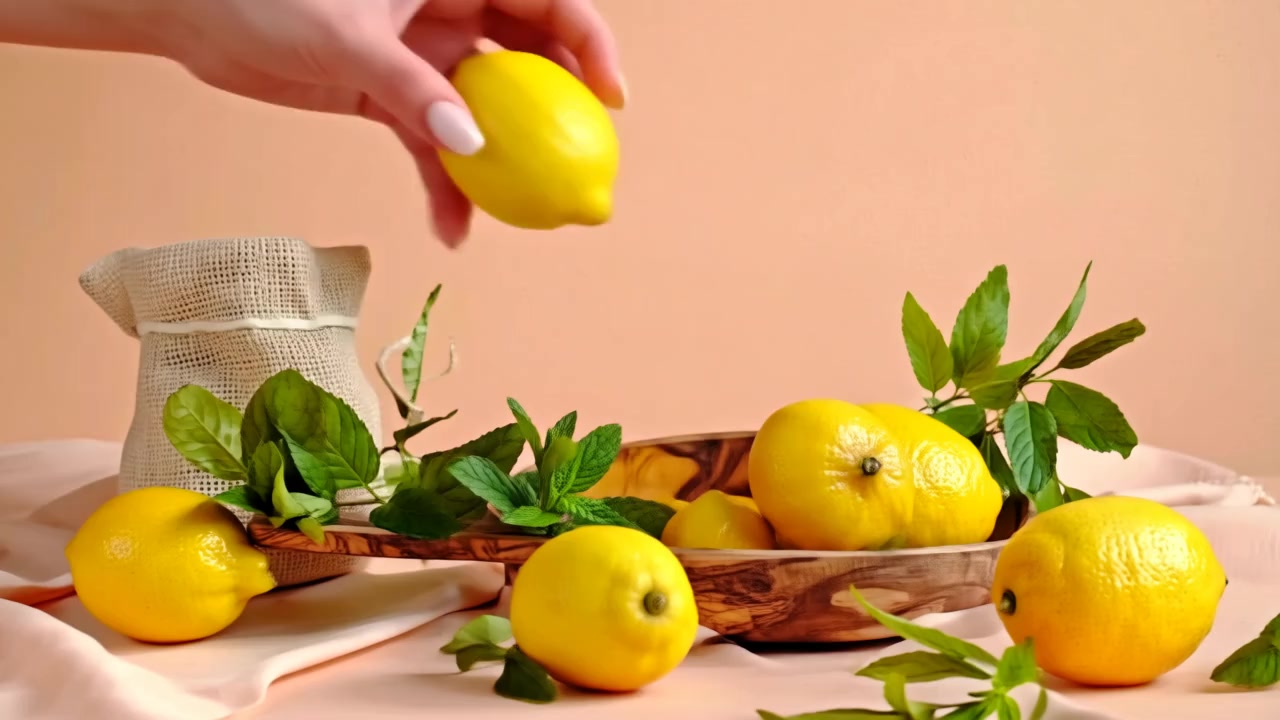} &
        \includegraphics[width=\linewidth]{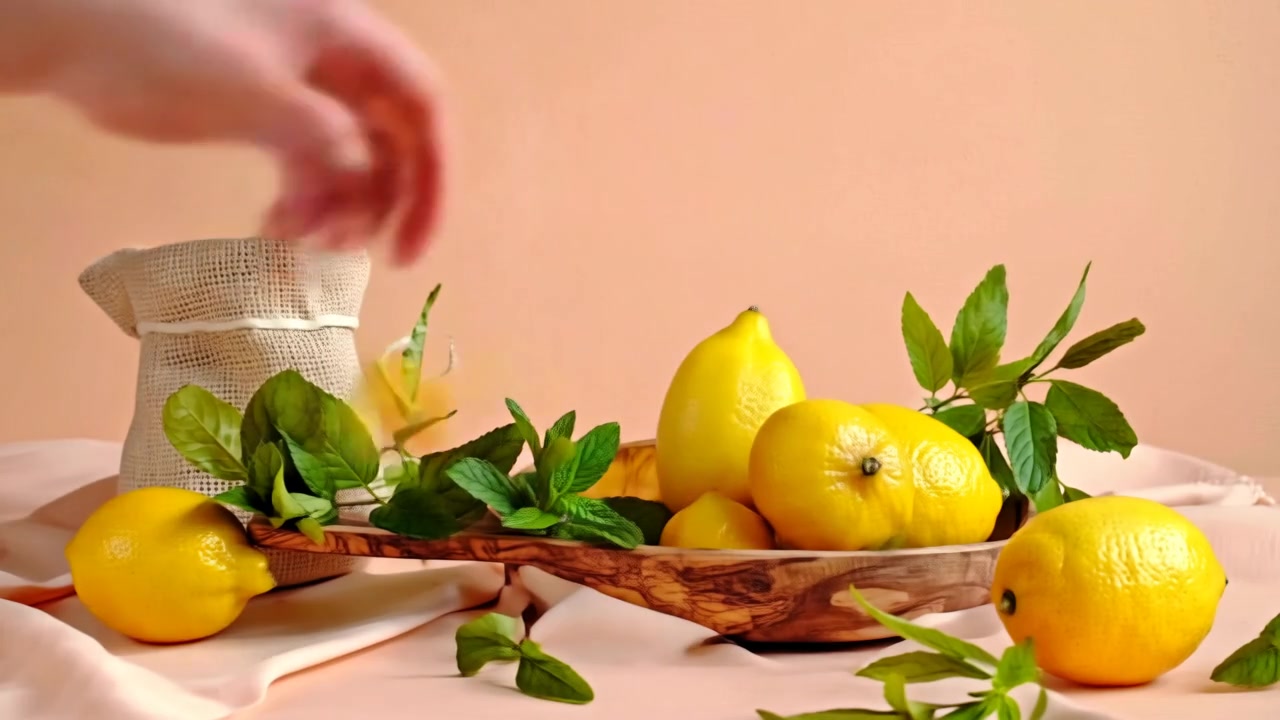} &
        \includegraphics[width=\linewidth]{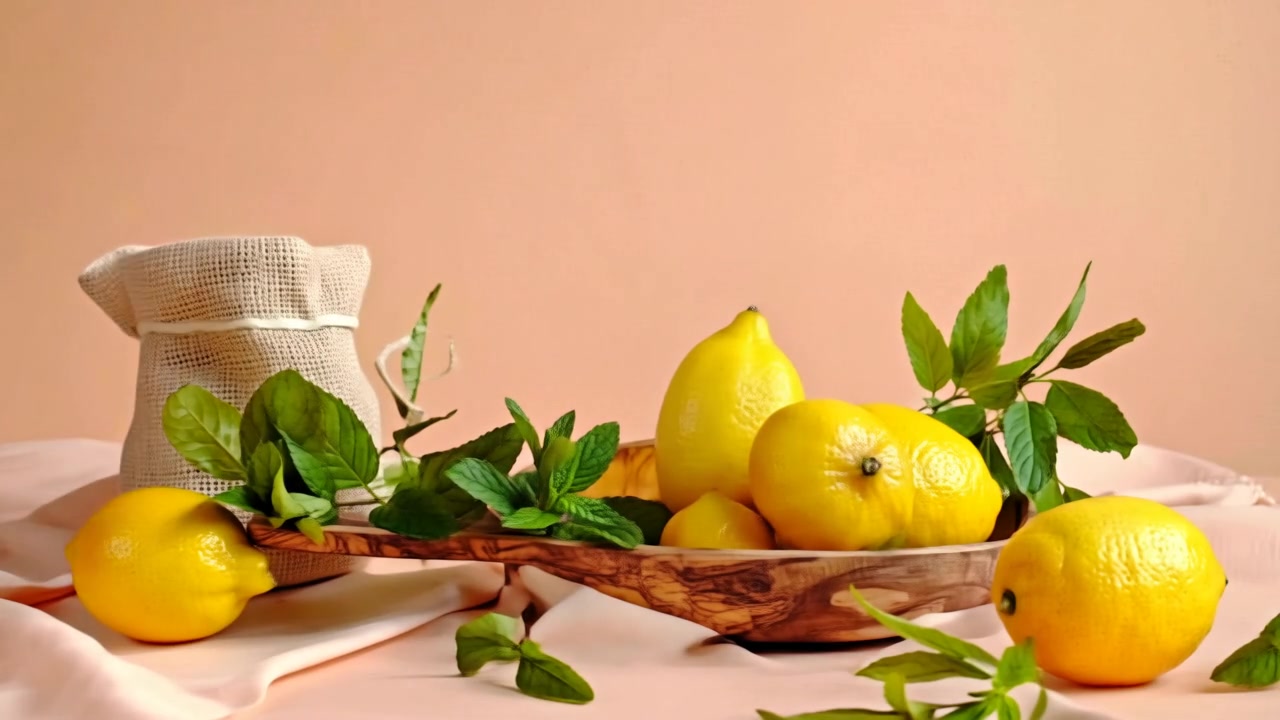} &
        
        {\animategraphics[width=\linewidth]{8}
        {figs/t2v_animates/vdmini_hunyuan/A_hand_with_delicate_fingers_picks_up_a_bright_yellow_lemon_from_a_wooden_bowl_filled_with_lemons_an/}{001}{125}}
  \end{tabular}
  \vspace{-1em}
  \caption{Qualitative Results of VDMini-T2V. For the first two rows, the \textbf{Prompt} is \underline{An astronaut flying in space, surrealism style.} The first row shows the video generated by T2V-Turbo-v2, while the second row presents the video generated by VDMini-T2V-Turbo. For the last two rows, the \textbf{Prompt} is \underline{A hand with delicate fingers \ldots}. The third row displays the video generated by HunyuanVideo, and the fourth row shows the video generated by VDMini-T2V-HY based on HunyuanVideo. From left to right: the 1st frame, the 10th frame, and the video are shown. \red{\animationNotes}.}
  \Description{qualitative results of VDMini-T2V}
  \label{fig:t2v_turbo_animate}
  \vspace{-1em}
\end{figure*}

%% file: tables/table_i2v_ablation.tex
\begin{table}[htbp]
  \vspace{-0.5em}
  \centering
  \begin{minipage}[htbp]{0.234\textwidth}
    \centering
    \caption{Effectiveness of loss components. Individual Content Distillation Loss ($\mathcal{L}_{ICD}$) and Integrated Content Adversarial Loss ($\mathcal{L}_{ICA}$) in the VDMini-I2V model.}
  \vspace{-1em}
    \label{tab:i2v_ablation_loss}
    \resizebox{\columnwidth}{!}{
      \begin{tabular}{ccc}
      \toprule
      $\mathcal{L}_{ICD}$ & $\mathcal{L}_{MCA}$ & \textbf{FVD} $\downarrow$ \\
      \hline
      $\times$ & $\times$ & 299.44 \\
      $\checkmark$ & $\times$ & 224.24 \\
      $\times$ & $\checkmark$ & 257.99 \\
      $\checkmark$ & $\checkmark$ & 198.13 \\
      \bottomrule
      \end{tabular}
    }
  \end{minipage}
  \hspace{0.4em}
  \begin{minipage}[htbp]{0.22\textwidth}
    \centering
    \caption{\raggedright Sensitivity analysis of hyperparameters $\lambda_{ICD}$ and $\lambda_{ICA}$, representing the loss weights for $\mathcal{L}_{ICD}$ and $\mathcal{L}_{ICA}$ in the VDMini-I2V model.}
  \vspace{-1em}
    \label{tab:i2v_ablation_lambda}
    \resizebox{\columnwidth}{!}{
      \begin{tabular}{ccc}
      \toprule
        $\lambda_{ICD}$ & $\lambda_{MCA}$ & \textbf{FVD} $\downarrow$  \\
        \hline
        1 & 1 & 254.30 \\
        0.01 & 1 & 286.42 \\
        0.1 & 0.5 & 198.63 \\
        0.1 & 1 & 198.13 \\
        \bottomrule
      \end{tabular}
    }
  \end{minipage}
  % \caption{Ablation study results for the VDMini-I2V model, showing the impact of (a) different loss components and (b) hyperparameters on the FVD metric.}
  \vspace{-1em}
\end{table}

%% file: tables/table_t2v_comparison.tex
\begin{table}[htbp]
  \vspace{-0.5em}
  \centering
  \caption{
    Comparison of VDMini-T2V with other methods on VBench-T2V in terms of Quality Score, Semantic Score, Total Score, and Latency.
  }
  \vspace{-1em}
  \label{tab:t2v_comparison}
  \scalebox{0.84}{
      \begin{tabular}{lccccc}
      \toprule
      \textbf{Methods} & \textbf{Quality}$\uparrow$ & \textbf{Semantic}$\uparrow$ & \textbf{Total}$\uparrow$  & \textbf{Latency (ms)}$\downarrow$\\
      \hline
      \textbf{UNet-based Model} &&&& \\
      \midrule
      Kling & 83.39 & 75.68 & 81.85 & - \\
      VideoCrafter2 & 82.20 & 73.42 & 80.44 & - \\
      T2V-Turbo-v2 & 85.13 & 77.12 & 83.52 & 2554.05 \\
      VDMini-T2V-Turbo & 83.33 & 77.38 & 82.14 & 1662.26 \\
      \midrule
      \textbf{DiT-based Model} &&&& \\
      \midrule
      FastVideo & 85.09 & 75.82 & 83.24 & 366.11 \\
      VDMini-T2V-HY & 84.34 & 74.76 & 82.42 & 292.89\\
      \bottomrule
      \end{tabular}
  }
  \vspace{-1em}
\end{table}

%% file: 10_conclusion.tex
\section{Conclusion}
\label{sec:conclusion}
In this work, we have presented VDMini, a lightweight Video Diffusion Models (VDMs). Our approach concentrates on reducing the inference time while maintaining the individual content and multi-frame content quality. The pruning strategy is designed by analyzing the importance of the blocks in the U-Net as well as the visual quality of the generated video, leading to the removal of redundant shallower layers and preserving the deeper layers to keep the quality of multi-frame content. To further enhance the motion dynamics and quality of the generated video, we apply a multi-frame content adversarial loss with individual content distillation loss during the fine-tuning process. We achieve an average 2.5 $\times$, 1.4 $\times$, and 1.25 $\times$ speed up for the I2V method SF-V, the T2V method T2V-Turbo-v2, and the T2V method HunyuanVideo, respectively.

% \nbf{Limitations and Future Work.} Our method is tailored for diffusion-based VDMs like SF-V, T2V-Turbo-v2, and HunyuanVideo, and is not directly applicable to auto-regressive (AR) models. Future work will focus on extending our approach to optimize AR-based video generation models.

% \nbf{Limitations and Future Work.} Our method is specifically designed for video diffusion models (VDMs) that rely on the diffusion process, such as the UNet-based models SF-V and T2V-Turbo-v2, and DiT-based models HunyuanVideo. As such, it icable to auto-regressive (AR) video generation models. In future work, we aim to extend our approach to optimize and accelerate AR-based video generation models.
% Additionally, token pruning~\cite{chen2022lsvc,chen2023neural} could be explored to further accelerate VDMs. 

%% file: 12_appendix.tex
\renewcommand{\thefigure}{A\arabic{figure}}
\renewcommand{\thetable}{A\arabic{table}}
\setcounter{section}{0} % Reset section counter
\setcounter{figure}{0} % Reset figure counter
\setcounter{table}{0}  % Reset table counter

\noindent This appendix provides details about Section~\ref{sec:appendix:vae-decoder}, the compression of the VAE decoder of SF-V; Section~\ref{sec:appendix:details-about-the-compressed-unet}, the compressed U-Net; Section~\ref{sec:appendix:details-sf-v-t2v-turbo-v2}, I2V and T2V baselines; Section~\ref{sec:appendix:hunyuanvideo}, the extensive experiments on the DiT-based HunyuanVideo.

\section{Compression of the VAE Decoder of SF-V}
\label{sec:appendix:vae-decoder}
The VAE decoder in the I2V method SF-V incurs significant inference time. Therefore, we follow~\cite{liSnapFusionTextImageDiffusion2023} to apply both layer pruning and channel pruning techniques~\cite{fang2023depgraph} to compress the VAE decoder, which will further reduce the inference time for SF-V.
Initially, we present the block-wise architecture of the VAE decoder in Figure~\ref{fig:supp:vae-decoder-architecture} and the block-wise inference time and the number of parameters in Figure~\ref{fig:supp:vae_blockwise_time_params}. Our observations indicate that the latter layers of the VAE decoder, which handle higher resolution features, consume most of the inference time.
Specifically, we remove layers in the ``MidBlock'' and ``UpBlock.3'', as these layers significantly contribute to inference time and model weight. Subsequently, we utilize Torch-Pruning~\cite{fang2023depgraph} for channel pruning to further accelerate the model. This results in a compressed VAE decoder with a 70\% reduction in inference time and a 30\% reduction in the number of parameters. By fine-tuning the compressed VAE decoder on the OpenVid-1M dataset~\cite{nanOpenVid1MLargeScaleHighQuality2024} using reconstruction loss, perceptual loss, and adversarial loss, we achieve performance comparable to the original VAE decoder.

\input{figs/supp_vae_architecture.tex}
\input{figs/supp_vae_blockwise_time_params.tex}

\noindent\textbf{Reconstruction Results of the Compressed VAE Decoder.}
The VAE decoder in the I2V task consumes a significant amount of inference time. To address this, we further compress the VAE decoder in the VDMini-I2V model. We employ layer pruning and channel pruning techniques to accelerate the VAE decoder, followed by fine-tuning on the OpenVid-1M dataset using reconstruction loss, perceptual loss, and adversarial loss as described in~\cite{esser2021taming}. As shown in Table~\ref{tab:model comparison}, the inference time of the compressed VAE decoder is reduced by 70\%, and the model has 30\% fewer parameters compared to the original. As provided in Table~\ref{tab:vae_decoder}, we report the PSNR, SSIM, and LPIPS on the UCF101 dataset, and the reconstruction results are illustrated in Figure~\ref{fig:i2v_vae_decoder}.

\input{figs/i2v_vae_decoder.tex}

\input{tables/table_model_comparison.tex}
\input{tables/table_i2v_decoder.tex}

\input{tables/table_t2v_model_comparison.tex}
\input{figs/supp_unet_blockwise_fvd.tex}
\input{figs/supp_unet_blockwise_time_params.tex}

\section{Details about the Compressed U-Net}
\label{sec:appendix:details-about-the-compressed-unet}
Most I2V and T2V models share nearly identical architectures in their diffusion model backbone (\textit{i.e.,} U-Net). For instance, classical VDMs such as VideoCrafter~\cite{chenVideoCrafter1OpenDiffusion2023} and Open-Sora~\cite{opensora} utilize the exact same U-Net architecture for both I2V and T2V tasks. Consequently, it is reasonable to assume that specific blocks within these U-Nets perform similar roles across both tasks.
Hence, in this work, we focus on providing an in-depth analysis of the I2V model SF-V and extend our observations to the T2V model T2V-Turbo-v2.

\begin{figure}
  \centering
  \scalebox{0.75}{
  \includegraphics{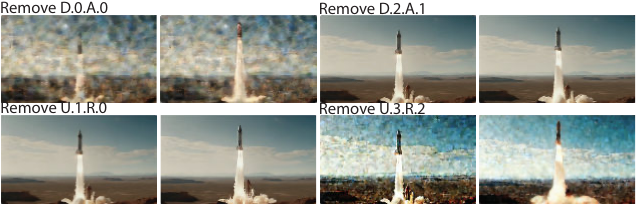}
  }
  \caption{Visual analysis of the pruned SF-V.}
  \Description{}
  \label{fig:visual_analysis_sf-v}
\end{figure}

Specifically, we study the \textbf{importance score} and the \textbf{computational complexity} for the U-Net of SF-V. 
First, we present the detailed block-wise importance scores by replacing the blocks with identity mapping or a single convolutional layer, as shown in Figure~\ref{fig:supp:unet_blockwise_fvd}. A higher FVD score indicates a more important block. The FVD score calculation follows the evaluation protocol introduced in StyleGAN-V~\cite{skorokhodov2022stylegan}. For fast evaluation, we sample 1200 videos from the UCF101 dataset as the ground-truth videos and calculate the FVD score between the generated and ground-truth videos. As illustrated in Figure~\ref{fig:supp:unet_blockwise_fvd}, we use a specific notation to denote the blocks in the U-Net. For example, the module named ``D.1.R.0.RB-S'' indicates: ``D'' for DownBlock, ``1'' for the first subblock, ``R'' for ResBlock type, ``0'' for the layer index of the ResBlock, and ``RB-S'' for SpatialResBlock consisting of 2D ResNet blocks. The results show that the SpatialResBlock in the DownBlocks is more important than other sub-blocks, and in the UpBlocks, the SpatialAttentionBlock is more important.
Second, we present the computational complexity study, including the inference time and the number of parameters for each block in the U-Net of SF-V, as shown in Figure~\ref{fig:supp:unet_blockwise_time_params}. The inference time measurement is conducted on a single NVIDIA A100 GPU. Additionally, we present the first and last frames of the videos generated by the pruned model in Figure~\ref{fig:visual_analysis_sf-v}. When removing D.2 or U.1, the generated videos exhibit subtle motion.

According to the above observations, we propose the following pruning strategy for VDMini. 
1) completely remove the ``Down-3'', ``Mid'', and ``Up-0'' blocks in VDMini, 
and 2) reduce the number of blocks in ``Down-0'', ``Down-1'', ``Up-2'', and ``Up-3'' by one compared to the original U-Net.

As explained previously, we directly apply our observations regarding U-Net compression from the I2V model to the T2V model. Consequently, as shown in Table~\ref{tab:supp:vdmini_compressed_unet}, the pruned models VDMini-T2V and VDMini-I2V share an almost identical architecture, with only minor differences in the structure of the "ResBlock." Specifically, in VDMini-T2V, the ResBlock is a 2D ResNet block, while in VDMini-I2V, it consists of both a 2D ResBlock and a 3D ResBlock.

\input{tables/supp_table_vdmini_compressed_unet.tex}
\input{tables/table_dit.tex}

\section{Implementation of I2V and T2V Baselines }
\label{sec:appendix:details-sf-v-t2v-turbo-v2}
Here, we introduce the implementation details of our I2V and T2V baseline methods SF-V and T2V-Turbo-v2.

\nbf{SF-V}:~\cite{zhangSFVSingleForward2024} is a one-step I2V model fine-tuned from Stable Video Diffusion (SVD)~\cite{blattmannStableVideoDiffusion2023} on a 1M internal video dataset. SF-V follows LADD~\cite{sauer2024fast} to fine-tune the pre-trained SVD with adversarial loss, with the discriminator initialized from the pre-trained SVD. Since the code and model of SVD are not publicly available, we re-implemented the SF-V model using the OpenVid-1M dataset, a high-quality video dataset with dense captions. We adhered to the original SF-V training settings, with a batch size of 32 and a gradient accumulation step of 4. The learning rates for the U-Net and the heads of the discriminator were set to 1e-5 and 1e-4, respectively. After 50K training steps, we achieved an FVD score of 166.26, comparable to the original SF-V model.

\nbf{T2V-Turbo-v2}~\cite{liT2VTurbov2EnhancingVideo2024} is a T2V model, which is
distilled from VideoCr-after2~\cite{chenVideoCrafter2OvercomingData2024} and follows a consistency distillation scheme. The initial version of T2V-Turbo (\ie T2V-Turbo-v1) adopts mixed reward models to enhance the generation quality and prompt consistency, and T2V-Turbo-v2 further employs MotionClone~\cite{lingMotionCloneTrainingFreeMotion2024} to enhance the motion dynamics in the few-step sampling, and the resulting model achieves the first rank in the VBench leaderboard. The U-Net architecture of T2V-Turbo-v2 is similar to SVD.

\nbf{HunyuanVideo} is a large-scale T2V model with cutting-edge performance. HunyuanVideo is based on the Diffusion Transformer (DiT) architecture, which is with size of 13B parameters in total. The model is consisted of multiple components: CLIP and MLLM for text encoding, a 3D VAE for image/video encoding and decoding, and a DiT for video latent generation. Inspired by SD3~\cite{esser2024scaling} HunyuanVideo is trained with a curated dataset with vase image/video-text pairs, the training framework follows the flow matching. The released model could achieve competitive generation quality with the state-of-the-art commercial models such as Luma and Gen-3.

\section{Implementation Details for VDMini-T2V-HY}
\label{sec:appendix:hunyuanvideo}
We acknowledge that the issues of subtle motion and short duration are inherent limitations of the base models. Therefore, we conduct extensive experiments on the DiT-based model HunyuanVideo~\cite{kong2024hunyuanvideo}, the results are reported in Table~\ref{tab:hunyuanvideo}.

For training the VDMini-T2V model based on HunyuanVideo, we follow the full fine-tuning strategy introduced in FastVideo, which includes velocity prediction loss introduced in flow matching and our proposed ICMD loss. The learning rate for DiT and the discriminator are set to 5e-6 and 1e-5, respectively. The batch size is set to 8 for fine-tuning with 30K steps. The loss weight for ICD and MCA loss are same as VDMini-T2V based on T2V-Turbo-v2.

%% file: figs/supp_vae_architecture.tex
% Use figure* for multi-column figure
\begin{figure}[thbp]
    \centering
    \includegraphics[width=\linewidth]{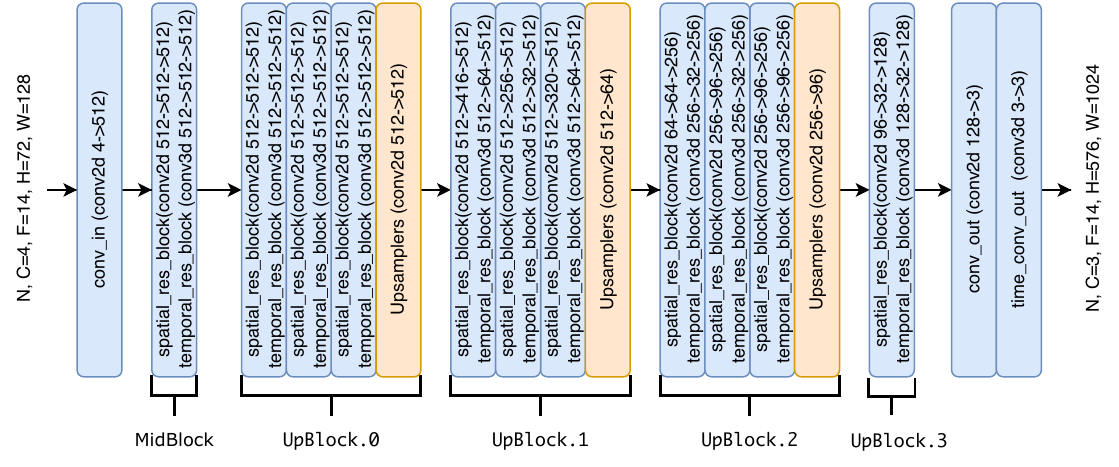}
    \caption{Network architecture of the compressed VAE decoder. First, we remove the layers in MidBlock and UpBlock.3, then perform channel pruning to speed up inference time.}
    \Description{network architecture of VAE decoder}
    \label{fig:supp:vae-decoder-architecture}
\end{figure}

%% file: figs/supp_vae_blockwise_time_params.tex
% Use figure* for multi-column figure
\begin{figure}[htbp]
    \centering
    \includegraphics[width=\linewidth]{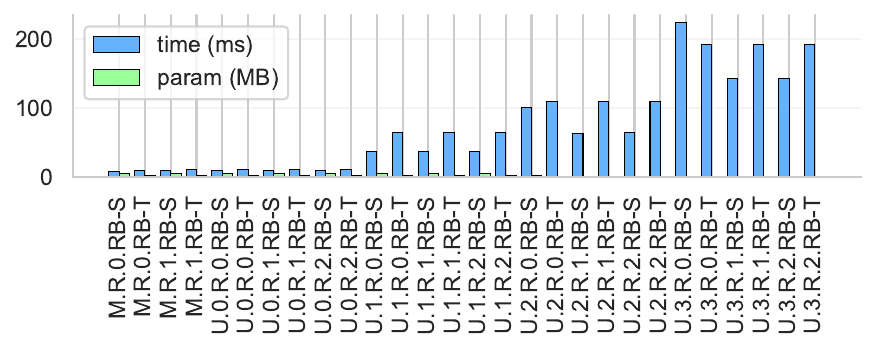}
    \caption{Time and Parameters of the blocks in the VAE decoder.}
    \Description{Time and Param Visualization}
    \label{fig:supp:vae_blockwise_time_params}
    \vspace{-1em}
\end{figure}

%% file: figs/i2v_vae_decoder.tex
% Use figure* for multi-column figure
\begin{figure}[htbp]
  \centering
  \includegraphics[width=\linewidth]{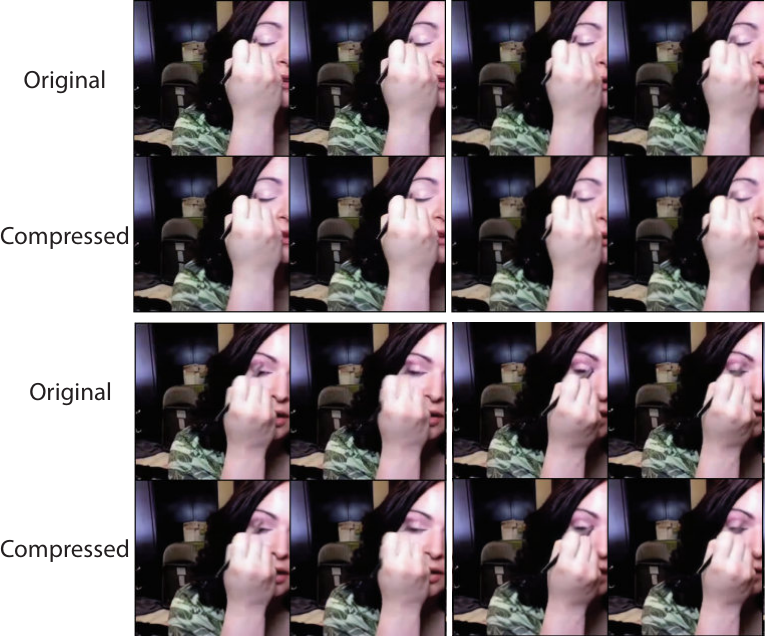}
  \caption{Reconstruction results of the VAE decoder. The compressed VAE decoder achieves image quality metrics comparable to the original VAE decoder while being significantly faster. For instance, decoding a video latent of size $14 \times 72 \times 128$ takes 2832ms with the original VAE decoder, compared to only 840.5ms with the compressed VAE decoder.}
  \Description{VAE decoder reconstruction}
  \vspace{-0.5em}
  \label{fig:i2v_vae_decoder}
\end{figure}

%% file: tables/table_model_comparison.tex
\begin{table}[htbp]
  % \vspace{-1em}
  \centering
  \caption{Comparison of the model architecture and inference latency between SF-V and our proposed VDMini-I2V. }
  \label{tab:model comparison}
  \resizebox{\columnwidth}{!}{
    \centering
    \begin{tabular}{cccccc}
    \toprule
    \multirow{2}{*}{Model} &                   & CLIP Encoder & VAE Encoder      & U-Net           & VAE Decoder     \\ \cline{2-6} 
                            & Resolution  & 224 $\times$ 224    & 576 $\times$ 1024      & 14 $\times$72 $\times$ 128   & 14 $\times$ 72$\times$ 128   \\ \hline
    \multirow{2}{*}{SF-V}  & \#Param (M)  & 632.08       & 34.16           & 1524.62         & 63.58           \\ 
                            & Latency (ms)      & 35           & 75              & 512             & 2832            \\ \hline
    \multirow{2}{*}{VDMini-I2V} & \#Param (M)  & 632.08       & 34.16           & \textbf{940.9}  & \textbf{39.17}  \\ 
                            & Latency (ms)      & 35           & 75              & \textbf{345}    & \textbf{840.5}  \\
    \bottomrule
    \end{tabular}
  }
  % \vspace{-1em}
\end{table}

%% file: tables/table_i2v_decoder.tex
\begin{table}[htbp]
  \centering
  \caption{
    Quantitative comparison of the image quality metrics on the UCF101 dataset between the original VAE decoder and the compressed VAE decoder.
  }
  \label{tab:vae_decoder}
  \resizebox{0.8\columnwidth}{!}{
    \begin{tabular}{l|ccc}
    \toprule
    \textbf{VAE Decoder} & \textbf{PSNR} $\uparrow$ & \textbf{SSIM} $\uparrow$ & \textbf{LPIPS} $\downarrow$ \\
    \hline
    Original & 32.30 & 0.94 & 0.031 \\
    Compressed & 31.41 & 0.93 & 0.039 \\
    \bottomrule
    \end{tabular}
  }
  % \vspace{-1em}
\end{table}

%% file: tables/table_t2v_model_comparison.tex
\begin{table}[htbp]
  % \vspace{-1em}
  \centering
  \caption{Comparison of the model architecture and inference latency between T2V-Turbo-v2 and our proposed VDMini-T2V.}
  \label{tab:t2v_model_comparison}
  \resizebox{\columnwidth}{!}{
    \begin{tabular}{ccccc}
    \toprule
    \multirow{3}{*}{Model} & & Text Encoder & U-Net & VAE Decoder \\
    \cline{2-5}
     & Resolution & 77 Tokens & 16 $\times$ 40 $\times$ 64 & 16 $\times$ 40 $\times$ 64 \\\cline{2-5}
     & Inference Steps & 1 & 16 & 1 \\
    \hline
    \multirow{2}{*}{T2V-Turbo-v2} & \#Param (M) & 354.03 & 1413.65 & 49.49 \\
     & Latency (ms) & 13.57 & 2554.05 & 367.66 \\
    \hline
    \multirow{2}{*}{VDMini-T2V-Turbo} & \#Param (M) & 354.03 & \textbf{817.02} & 49.49 \\
     & Latency (ms) & 13.57 & \textbf{1662.26} & 367.66 \\
    \bottomrule
    \end{tabular}
    }
  % \vspace{-1em}
\end{table}

%% file: figs/supp_unet_blockwise_fvd.tex
% Use figure* for multi-column figure
\begin{figure*}[htbp]
    \centering
    \includegraphics[width=0.99\linewidth]{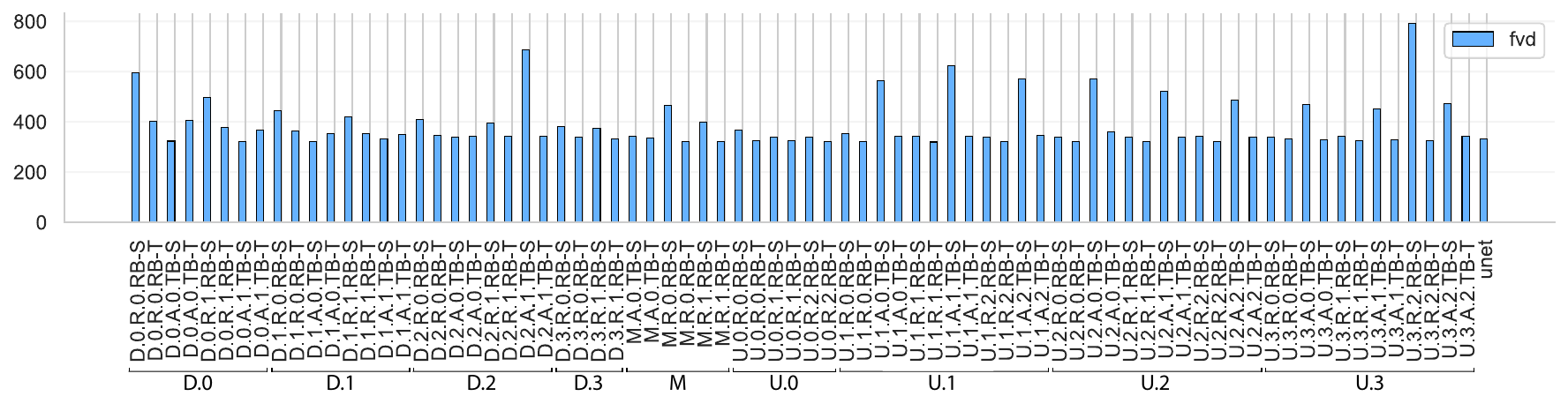}
    \caption{Block-wise FVD score by removing or replacing the blocks in the U-Net.}
    \Description{Block-wise FVD score}
    \label{fig:supp:unet_blockwise_fvd}
    \vspace{-0.5em}
\end{figure*}

%% file: figs/supp_unet_blockwise_time_params.tex
% Use figure* for multi-column figure
\begin{figure*}[htbp]
    \centering
    \includegraphics[width=0.99\linewidth]{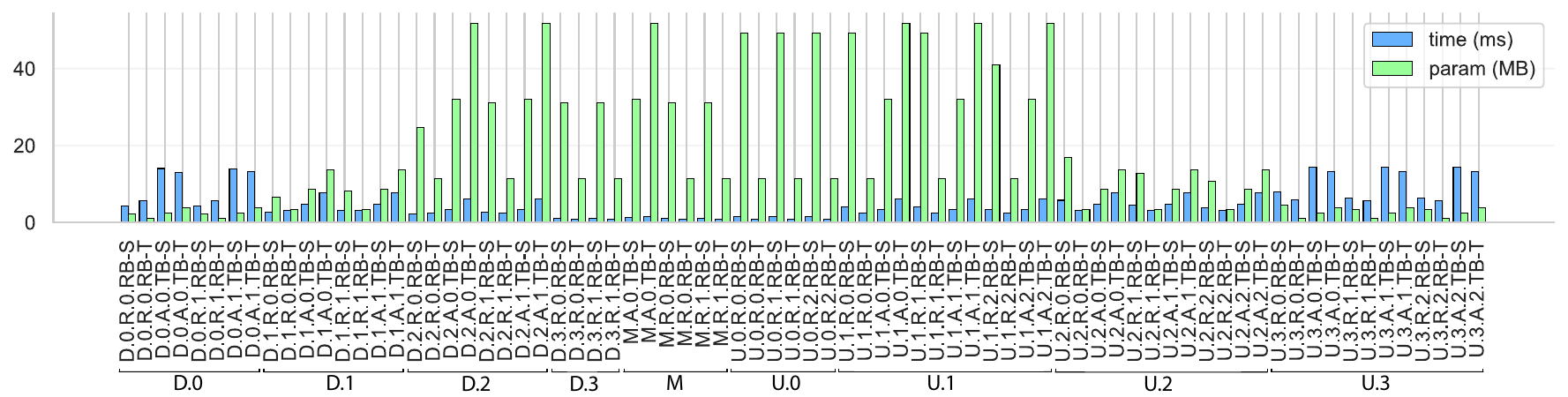}
    \caption{Block-wise inference time and parameter count in the U-Net.}
    \Description{}
    \label{fig:supp:unet_blockwise_time_params}
    \vspace{-0.5em}
\end{figure*}

%% file: tables/supp_table_vdmini_compressed_unet.tex
\begin{table}[ht]
\centering
  \caption{The detailed architecture of the compressed U-Net used in VDMini.}
  \label{tab:supp:vdmini_compressed_unet}
  \resizebox{0.99\columnwidth}{!}{
    \begin{tabular}{c|c|c|c|c|cc}
    \toprule
    \hline
    \multirow{2}{*}{\textbf{Stage}} & \multirow{2}{*}{\textbf{Resolution}} & \multirow{2}{*}{\textbf{Type}} & \multirow{2}{*}{\textbf{Config}} & \multicolumn{2}{c}{\textbf{UNet Model}} \\ \cline{5-6}
       & & & & \textbf{Origin} & \textbf{Ours} \\ \hline
      \multirow{4}{*}{Down-0} & \multirow{4}{*}{$T \times \frac{H}{8} \times \frac{W}{8}$} & \multirow{2}{*}{ResBlock} & Dimension & 320 & 320 \\
       & & & \# Blocks & 2 & 1 \\\cline{3-6}
       & & \multirow{2}{*}{TransformerBlock} & Dimension & 320 & 320 \\
       & & & \# Blocks & 2 & 1 \\ \hline
       \multirow{4}{*}{Down-1} & \multirow{4}{*}{$T \times \frac{H}{16} \times \frac{W}{16}$} & \multirow{2}{*}{ResBlock} & Dimension & 640 & 640 \\
       & & & \# Blocks & 2 & 1 \\\cline{3-6}
       & & \multirow{2}{*}{TransformerBlock} & Dimension & 640 & 640 \\
       & & & \# Blocks & 2 & 1 \\ \hline
       \multirow{4}{*}{Down-2} & \multirow{4}{*}{$T \times \frac{H}{32} \times \frac{W}{32}$} & \multirow{2}{*}{ResBlock} & Dimension & 1280 & 1280 \\
       & & & \# Blocks & 2 & 2 \\\cline{3-6}
       & & \multirow{2}{*}{TransformerBlock} & Dimension & 1280 & 1280 \\
       & & & \# Blocks & 2 & 2 \\ \hline
       \multirow{2}{*}{Down-3} & \multirow{2}{*}{$T \times \frac{H}{64} \times \frac{W}{64}$} & \multirow{2}{*}{ResBlock} & Dimension & 1280 & 1280 \\
       & & & \# Blocks & 2 & 0 \\ \hline
       \multirow{4}{*}{Mid} & \multirow{4}{*}{$T \times \frac{H}{64} \times \frac{W}{64}$} & \multirow{2}{*}{ResBlock} & Dimension & 1280 & 1280 \\
       & & & \# Blocks & 2 & 0 \\\cline{3-6}
       & & \multirow{2}{*}{TransformerBlock} & Dimension & 1280 & 1280 \\
       & & & \# Blocks & 1 & 0 \\ \hline
       \multirow{2}{*}{Up-0} & \multirow{2}{*}{$T \times \frac{H}{64} \times \frac{W}{64}$} & \multirow{2}{*}{ResBlock} & Dimension & 1280 & 1280 \\
       & & & \# Blocks & 3 & 0 \\ \hline
      \multirow{4}{*}{Up-1} & \multirow{4}{*}{$T \times \frac{H}{32} \times \frac{W}{32}$} & \multirow{2}{*}{ResBlock} & Dimension & 1280 & 1280 \\
       & & & \# Blocks & 3 & 3 \\\cline{3-6}
       & & \multirow{2}{*}{TransformerBlock} & Dimension & 1280 & 1280 \\
       & & & \# Blocks & 3 & 3 \\ \hline
       \multirow{4}{*}{Up-2} & \multirow{4}{*}{$T \times \frac{H}{16} \times \frac{W}{16}$} & \multirow{2}{*}{ResBlock} & Dimension & 640 & 640 \\
       & & & \# Blocks & 3 & 2 \\\cline{3-6}
       & & \multirow{2}{*}{TransformerBlock} & Dimension & 640 & 640 \\
       & & & \# Blocks & 3 & 2 \\ \hline
       \multirow{4}{*}{Up-3} & \multirow{4}{*}{$T \times \frac{H}{8} \times \frac{W}{8}$} & \multirow{2}{*}{ResBlock} & Dimension & 320 & 320 \\
       & & & \# Blocks & 3 & 2 \\\cline{3-6}
       & & \multirow{2}{*}{TransformerBlock} & Dimension & 320 & 320 \\
       & & & \# Blocks & 3 & 2 \\ \hline
    \bottomrule
    \end{tabular}
    }
  % \vspace{-1em}
\end{table}

%% file: tables/table_dit.tex
\begin{table*}[ht]
  % \vspace{-1em}
  \centering
  \caption{Evaluation of Hunyuan model on VBench-T2V.}
  \label{tab:hunyuanvideo}
  \scalebox{0.85}{
  \begin{tabular}{l|c c c c c c}
    \toprule
    Models & \textbf{Quality Score} $\downarrow$ & \textbf{Semantic Score} $\downarrow$ & \textbf{Total Score} $\downarrow$ & $\#$\textbf{Double-Stream Layer} & $\#$\textbf{Single-Stream Layer} & \textbf{Inference Speed(ms)} \\
    \midrule
    HunyuanVideo & 85.09\% & 75.82\% & 83.24\% & 20 & 40 & 366.11 \\ % 366.10874125
    VDMini-T2V-HY & 84.34\% & 74.76\% & 82.42\% & 12 & 28 & 292.89 \\ % 292.886993
    \bottomrule
  \end{tabular}
  }
\end{table*}